%% file: PaperForReview.tex
\documentclass[10pt,twocolumn,letterpaper]{article}
\usepackage[final]{cvpr2024}      

\usepackage{multicol}
\usepackage{graphicx}
\usepackage[table,xcdraw]{xcolor}
\usepackage{graphicx}
\usepackage{amssymb}
\usepackage{booktabs}
\usepackage{bbding}

\usepackage{algorithmic}
\usepackage{algorithm}
\usepackage{pifont}
\usepackage[normalem]{ulem}
\definecolor{cvprblue}{rgb}{0.21,0.49,0.74}
\usepackage[pagebackref,breaklinks,colorlinks,citecolor=cvprblue]{hyperref}

\usepackage[capitalize]{cleveref}
\crefname{section}{Sec.}{Secs.}
\Crefname{section}{Section}{Sections}
\Crefname{table}{Table}{Tables}
\crefname{table}{Tab.}{Tabs.}

\def\paperID{7124} 

\def\confName{CVPR}
\def\confYear{2024}

\title{CGI-DM: Digital Copyright Authentication for Diffusion Models via Contrasting Gradient Inversion}

\author{Xiaoyu Wu\textsuperscript{1}, Yang Hua\textsuperscript{2}, Chumeng Liang\textsuperscript{3}, Jiaru Zhang\textsuperscript{1}, Hao Wang\textsuperscript{4}, Tao Song\textsuperscript{1}\footnote{Corresponding Author}, Haibing Guan\textsuperscript{1}\\
Shanghai Jiao Tong University\textsuperscript{1}, Queen’s University Belfast\textsuperscript{2}, \\University of Southern California\textsuperscript{3}, Louisiana State University\textsuperscript{4}\\
{\tt\small \{wuxiaoyu2000, jiaruzhang, songt333, hbguan\}@sjtu.edu.cn 
}
\and
{
\tt\small Y.Hua@qub.ac.uk , 
\tt\small chumengl@usc.edu, haowang@lsu.edu 
}
}

\begin{document}
\maketitle
\begin{abstract}
Diffusion Models (DMs) have evolved into advanced image generation tools, especially for few-shot generation where a pretrained model is fine-tuned on a small set of images to capture a specific style or object. Despite their success,  concerns exist about potential copyright violations stemming from the use of unauthorized data in this process. In response, we present Contrasting Gradient Inversion for Diffusion Models (CGI-DM), a novel method featuring vivid visual representations for digital copyright authentication. Our approach involves removing partial information of an image and recovering missing details by exploiting conceptual differences between the pretrained and fine-tuned models. We formulate the differences as KL divergence between latent variables of the two models when given the same input image, which can be maximized through Monte Carlo sampling and Projected Gradient Descent (PGD). The similarity between  original and recovered images serves as a strong indicator of potential infringements. Extensive experiments on the WikiArt and Dreambooth datasets demonstrate the high accuracy of CGI-DM in digital copyright authentication, surpassing alternative validation techniques.  Code  implementation is available at \url{https://github.com/Nicholas0228/Revelio}. \

\end{abstract}

\addtocounter{footnote}{1}\footnotetext{\noindent Correspondence to  Tao Song (songt333@sjtu.edu.cn).}

\input{inputs/introduction}

\input{inputs/background}

\input{inputs/method_abb}
\input{inputs/experiment}

\input{inputs/related_work}
\input{inputs/conclusion}
\input{inputs/acknowledgements}
\newpage
~~\\
\newpage
\appendix

\input{inputs/appendix}

{\small
\bibliographystyle{ieee_fullname}
\bibliography{main}
}

\end{document}

%% file: inputs/introduction.tex
\section{Introduction}
\label{sec:intro}
\begin{figure}[t]

\begin{center}
\includegraphics[width=0.95\columnwidth]{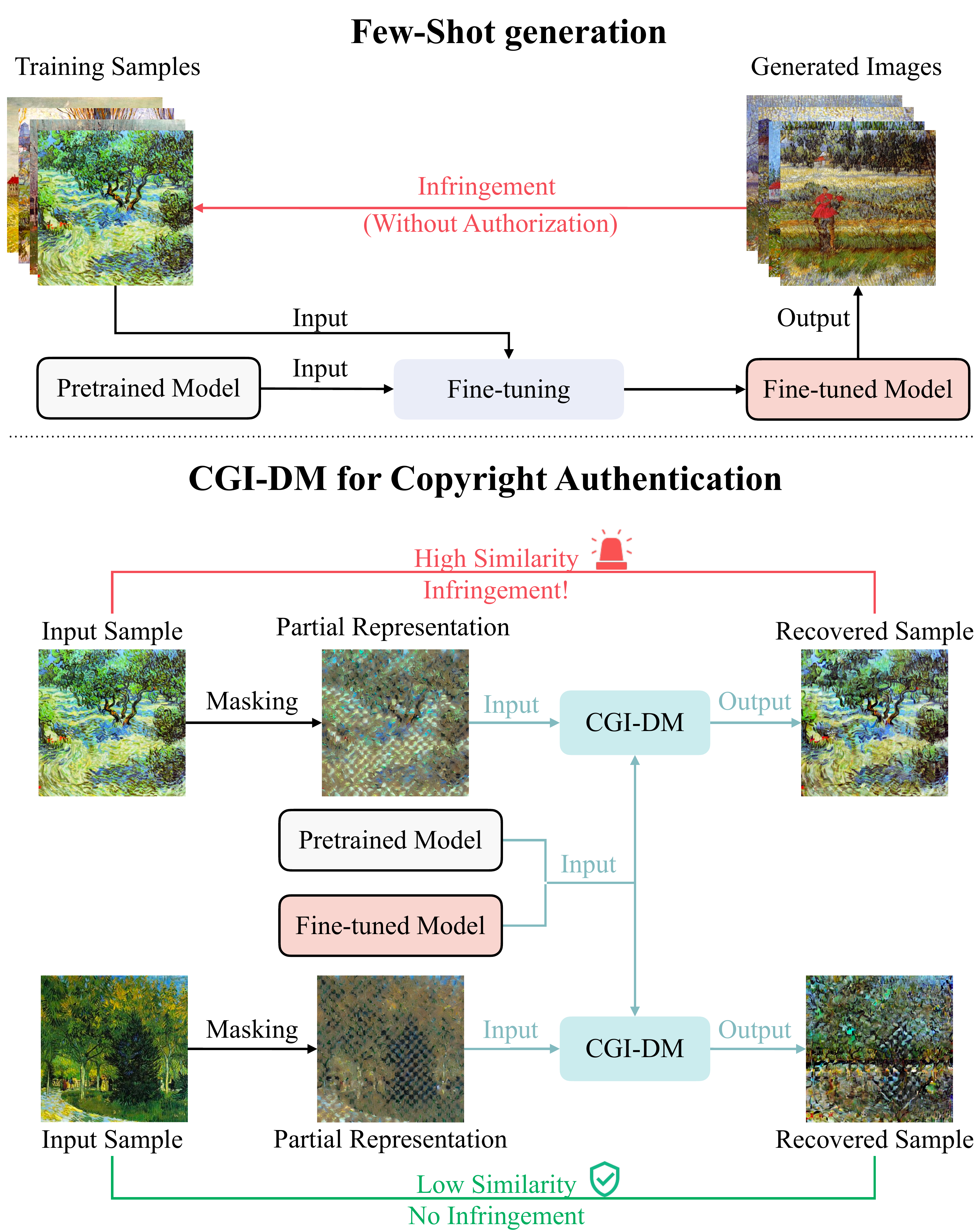}

    \caption{\textbf{Top:} Infringements brought by few-shot generation. A small set of images is used to fine-tuning the pretrained model. The fine-tuned model is then capable of high-quality generation, which, if performed without proper authorization, may lead to infringements. \textbf{Bottom:} Our method CGI-DM for copyright authentication. CGI-DM recovers the missing details from partial representation of an input sample. Then the the similarity between recovered samples and input samples can be used to validate infringements.}
    \label{fig:intuition}
\end{center}
\vspace{-0.35in}
\end{figure}

Recent years have witnessed the advancement of Diffusion Models (DMs) in computer vision. These models demonstrate exceptional capabilities across a diverse array of tasks, including image editing~\cite{kawar2022imagic}, and video editing~\cite{yang2022diffusion}, among others. Particularly, the emergence of few-shot generation techniques, exemplified by Dreambooth~\cite{dreambooth} and Lora~\cite{lora},  has considerably reduced the costs  associated with replicating artwork and transferring art styles, all the while maintaining an exceedingly high standard of quality. As illustrated in~\cref{fig:intuition} (Top), these methods focus on swiftly capturing the style or primary objects by fine-tuning a pretrained model on a small set of images. This process enables efficient and high-quality art imitation and art style transfer by utilizing the fine-tuned model.

However, these advanced few-shot generation techniques also spark widespread concerns regarding the protection of copyright for human artworks and individual photographs. There is a growing fear that parties may exploit the generative capabilities of DMs to create and profit from derivative artworks based on existing artists' works, without obtaining proper authorization~\cite{kimjungginews,mimicnews}. Concurrently, concerns arise regarding the creation of fabricated images of individuals without their consent~\cite{deepfake:dm}. All of these collectively pose a serious threat to the security of personal data and intellectual property rights.

To address these critical concerns, a line of approaches focuses on safeguarding individual images by incorporating adversarial attacks, such as AdvDM~\cite{advdm}, Glaze~\cite{glaze}, and Anti-Dreambooth~\cite{anti-dreambooth}. The adversarial attacks can disrupt the generative output, rendering the images unlearnable by diffusion models. These methods are implemented ahead of the fine-tuning process, and as such, we consider them as \textbf{precaution} approaches.

Another line of approaches facing such threats is copyright authentication. Copyright authentication compares the similarity between the output images  of diffusion models and the given images to validate unauthorized usage.  Such a process can serve as  legal proof for validating infringement (See~\cref{legal} for more details), and has been utilized as evidence in ongoing legal cases concerning violations enabled by DMs~\cite{legalnews}. This process happens after the fine-tuning and thus we consider it as a \textbf{post-caution} approach. However, current copyright authentication methods face difficulties in producing output images closely resembling training samples due to the pursuit of diversity in generative models. Consequently,  it becomes difficult to ascertain whether a particular training sample has been utilized solely based on the generated output of the model for post-caution methods.


In this paper, we propose a new  copyright authentication framework, named Contrasting Gradient Inversion for Diffusion Models (CGI-DM) to greatly improve the efficacy of the post-caution path, illustrated in~\cref{fig:intuition} (Bottom).  Recent advances in gradient inversion ~\cite{gi1, gi2, gi3} emphasize the importance of prior information in data extraction.   Inspired by this,  we propose first removing half of a given image. Then we utilize the retained partial representation as a prior and employ gradient inversion to reconstruct the original image.  As recent studies~\cite{mem1, mem2} indicate that generative models tend to ``memorize", the recovery of removed information should be possible only when the images are utilized during the fine-tuning process, enabling ``memorization" on given samples. Thus, a high similarity between the recovered image and the original image can indicate that the model has been trained with the given image.

 However, directly applying gradient inversion may not yield useful information for DMs, possibly because they inherently eliminate noise (see~\cref{direct-gi} for details).
To address this issue, we focus on contrasting two models: the  pretrained  and the fine-tuned model. Specifically, our goal is to leverage the conceptual differences between these two models. We  measure this disparity through the KL divergence between the latent variable distributions of the pretrained and fine-tuned models. Subsequently, we provide a proof demonstrating that maximizing this KL divergence approximates accentuating the loss differences between the two models. Building on this, we employ Monte Carlo Sampling on the noise and the time variable  during the diffusion process, utilizing PGD~\cite{pgd, advdm} to optimize the aforementioned loss difference. 
Comprehensive experiments are conducted on artists' works and objects, addressing the potential for unauthorized style transfers and the creation of fabricated images, both of which necessitate copyright authentication. The experiment results affirm the effectiveness and robustness of our approach.

In summary, our contributions are as follows:

\begin{itemize}

    \item We have formulated a novel post-caution approach for copyright protection—copyright authentication. This method complements precaution measures and provides robust legal proof of infringements.

    \item We propose  a new framework, CGI-DM,  for authentication. Utilizing Monte Carlo sampling and PGD optimization, we employ gradient inversion based on the partial representation of a given image.  The similarity between the recovered samples and original samples can serve as a robust and visible indicator for infringements. Notably, while most gradient inversion methods focus on classification models, we pioneer its application in the domain of generative models, emphasizing a new approach.

    \item We conduct extensive experiments on the WikiArt and Dreambooth datasets to substantiate the efficacy and robustness of our approach in distinguishing training samples from those not used during training. These demonstrate our method's effectiveness in authenticating digital copyrights and thus validating infringements for both style mimicry and fabricated images.
\end{itemize}

%% file: inputs/background.tex
\begin{figure*}[ht]
\vspace{-0.1in}
\begin{center}
\includegraphics[width=1.00\linewidth]{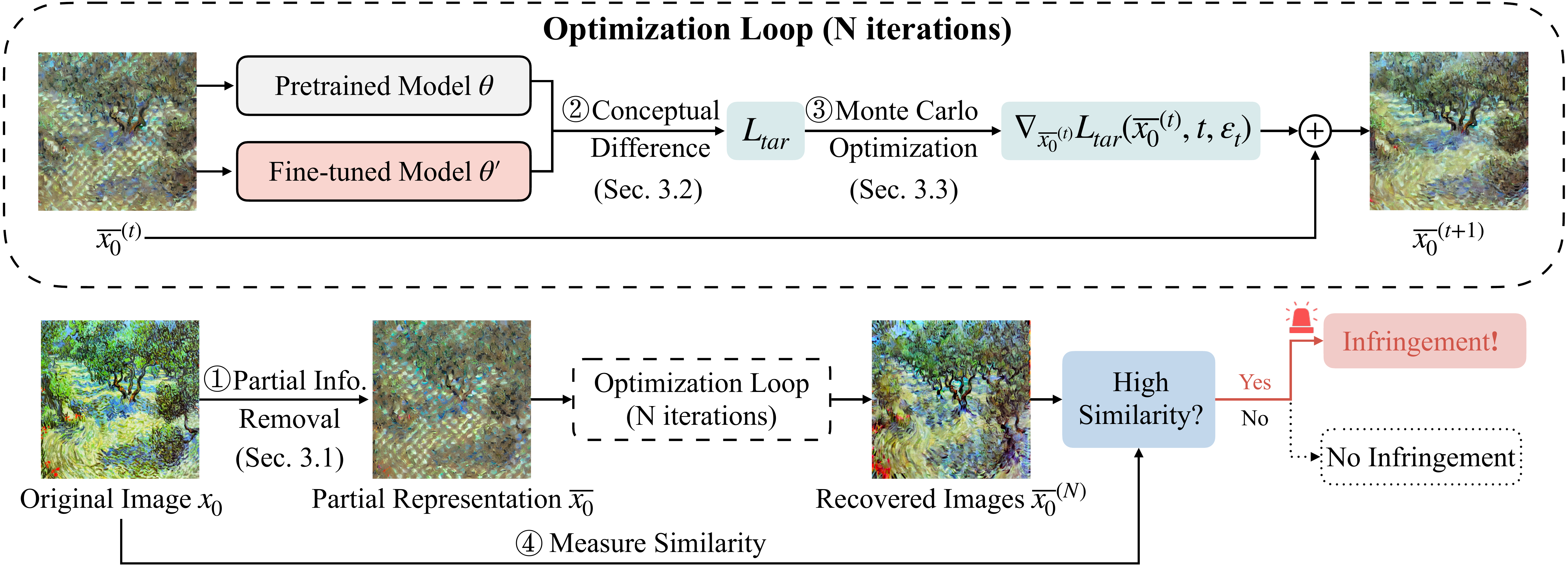}

\caption{Our framework, CGI-DM, for copyright authentication. We begin with a given image \(x_{0}\). \ding{172} We first remove part of \(x_{0}\), obtaining a partial representation \(\overline{x}_{0}\) of the image (refer to \cref{partial}). This partial representation is then fed into the optimization loop. \ding{173} Within the optimization loop, we leverage the conceptual disparity between the pretrained model \(\theta\) and the fine-tuned model \(\theta'\) given the partial representation \(\overline{x}_{0}\) (refer to \cref{kl}) to recover the missing details. Such disparity is formulated as $L_{tar}$. \ding{174} Subsequently, we employ Monte Carlo sampling on time variable $t$ and random noise $\varepsilon_{t}$ to optimize $L_{tar}$ (refer to \cref{alg}), getting step-wise gradient for updating the image. Over $\rm{N}$ steps of updating, the optimization loop produces the final recovered image $\overline{x}_{0}^{(N)}$. \ding{175} We authenticate copyright by measuring the similarity between recovered image \(\overline{x}_{0}^{(N)}\) with the original image \(x_{0}\).}
\label{fig:framework}
\end{center}
\vspace{-0.3in}
\end{figure*}

\section{Background}
\subsection{Diffusion Models}
\label{background}
Diffusion models are latent variable models that take the form of \( p_\theta(x_0) := \int p_\theta(x_{0:T} ) dx_{1:T} \), where \(x_{1:T}:= x_{1}, x_{2}, \cdots, x_{t}\) represents the latent variables of the same dimensionality as the data  $x_{0}$ belonging to real data distribution $q(x_{0})$: \(x_{0} \sim q(x_{0})\). The joint distribution of these latent variables is defined as a Markov chain:

\begin{equation}
    p_{\theta}(x_{0:T} ):= p_\theta(x_{T}) \prod_{t=1}^{N}p_{\theta}(x_{t-1}|x_{t}),
\end{equation}
where the transitions \(p_{\theta}(x_{t-1}|x_{t})\) are defined as learned Gaussian distributions: \(p_{\theta}(x_{t-1}|x_{t})=\mathcal{N}(x_{t-1}; \mu_{\theta}(x_{t}, t),\sigma_{t}^{2}\textbf{I})\). The initial step \(p_\theta(x_{T})\) of the Markov chain is defined as a normal distribution \(p_\theta(x_{T})=\mathcal{N}(x_{T};0, \textbf{I})\).

One of the most distinguishing properties of the diffusion model is that it leverages an approximate posterior \(q(x_{1:T} |x_{0})\) in the so-called diffusion process, which is also a Markov chain and can be simplified as: \(q(x_{t}|x_{0})=\mathcal{N}(x_{t}; \sqrt{\alpha_{t}}x_{0} , \sqrt{1-\alpha_{t}}\textbf{I})\). In other words, we can derive the latent variable \(x_{t}\) directly:

\begin{equation}
    \label{dm}
    x_{t} = \sqrt{\alpha_{t}}x_{0} + \sqrt{1-\alpha_{t}}\varepsilon_{t},
\end{equation}
where \(x_0\) is the original image, and \(\varepsilon_{t} \in \mathcal{N}(0,1)\) is the noise added at the current time $t$. Subsequently, in the so-called denoising process, a noise-prediction model $\epsilon_{\theta}$, parameterized by weights $\theta$, is employed to predict noise in the noisy image $x_{t}$. The predicted noise, denoted as $\epsilon_{\theta}(x_{t}, t)$, is then removed to recover a clear image from the noisy one.

The training target of the diffusion model begins with maximizing the aforementioned probability \(p_{\theta}(x_{0})\) for all data points \(x_{0}\) within the real data distribution. With a variational bound on the negative log-likelihood and a series of simplifications \cite{ho2020denoising}, the final objective can be transformed into the expectation of the mean squared error (MSE) loss on  the noise prediction error when the time variable $t$ and the noise $\varepsilon_{t}$ added are sampled:

\begin{equation}
    \label{target_dm}
        \theta = \arg\min\limits_{\theta}\mathbb{E}_{t, \varepsilon_{t} \sim \mathcal{N}(0,1)}\Vert\varepsilon_{t} - \epsilon_{\theta}(x_{t}, t)\Vert^{2}.\\
\end{equation}

\subsection{Copyright Authentication}
\label{background:copyright}

Copyright authentication aims to establish legal proof of infringements. By utilizing the pretrained model $\theta$ and a fine-tuned model $\theta'$ fine-tuning on a training dataset $\rm{X}$, the objective is to ascertain whether an image $x_0$ belongs to the training set  $\rm{X}$ in a way that is understandable to humans. When confirming inclusion in the training set,  indicating unauthorized usage, it is crucial to present a clear and vivid representation as legal evidence. Therefore, the method for copyright authentication  needs possess two essential properties:

\noindent\textbf{Accuracy.} The method should accurately classify samples used during training from those not used during training, preventing misleading outcomes.

\noindent\textbf{Explainability and Visualizability.} The method should be explainable and visualizable, especially for vision-generative models, ensuring its effectiveness as legal proof (refer to~\cref{legal} for more details). This implies that any metric used to distinguish between training and non-training samples should be in line with human vision, avoiding direct combination with the model's loss function, which might be incomprehensible to humans.

%% file: inputs/method_abb.tex
\section{Method}
\label{method-sec}
As depicted in~\cref{fig:framework}, our method, CGI-DM, is based on several processes: Initially, we remove part of the original image $x_0$, deriving partial representation $\overline{x_{0}}$ . Subsequently, we strive to recover the missing details by exploiting the conceptual differences between the pretrained model $\theta$ and the fine-tuned model $\theta'$. The process begins with maximizing the KL divergence between the probability distributions derived by the two models for latent variables of $\overline{x}_{0}$, the optimization problem of which can be solved with Monte Carlo Sampling. After optimization, the final output image $\overline{x}_0^{(N)}$ is then compared with the original image $x_0$. A high degree of similarity should be observed when the original image $x_0$ is used during the fine-tuning, while a significant discrepancy is expected when it is not.
\subsection{Removing Partial Information}
\label{partial}
The effectiveness of CGI-DM hinges significantly on the method employed to derive partial information $\overline{x}_{0}$. Recognizing that the quality and difficulty of recovering missing details depend on the type of information removed—be it detailed, background, or structural—we explore various techniques that consider these aspects during the partial information removal process (See~\cref{masked} in~\cref{sec:partial} for more details).


\subsection{Exploiting Conceptual Difference  }
\label{kl}

For a pretrained model $\theta$ and a fine-tuned model $\theta'$, we then aim to modify  partial representation $\overline{x}_0$ of the training example $x_0$ to best exploit the conceptual disparities between the two models. Such a process can recover the missing details in  $\overline{x}_0$ if the original sample $x_0$ is included in the training dataset $\rm{X}$.

We rely on the latent variable $\overline{x}_{1:T}$ of $\overline{x}_0$ and utilize the KL divergence between the probabilities of the latent variables $\overline{x}_{1:T}$ given $\overline{x}_0$ with respect to both models to measure such conceptual difference: $D_{\rm{KL}}(p_{\theta'}(\overline{x}_{1:T}'|\overline{x}_{0}')||{p_{\theta}(\overline{x}_{1:T}'|\overline{x}_{0}')})$. The process that gradually modifies change $\overline{x_{0}}$ to capture the conceptual difference between the two models can be formulated as optimizing the perturbation added to $\overline{x_{0}}$:




\begin{equation}
\label{eq:definition}
\begin{aligned}
\delta &:= \arg\max\limits_{\delta}D_{\rm{KL}}(p_{\theta'}(\overline{x}_{1:T}'|\overline{x}_{0}')||{p_{\theta}(\overline{x}_{1:T}'|\overline{x}_{0}')}), \\
    &\mbox{where } \overline{x}_{0}' = \overline{x}_0 + \delta, \Vert\delta\Vert\leq\epsilon, \epsilon\mbox{ is}\mbox{ a constant }.
\end{aligned}
\end{equation}

Leveraging the property of Markov chain and the training process of diffusion models, we show that such KL divergence  can be transformed into a closed form considering the difference between the latent variable and the mean value of the probability function in forward denoising diffusion implicit model (DDIM)~\cite{DDIM, diffusion-clip}:

\begin{equation}
    \begin{aligned}
  &\arg\max\limits_{\delta}\mathbb{E}_{p_{\theta'}(\overline{x}_{1:T}'|\overline{x}_{0}')}\log \frac{p_{\theta'}(\overline{x}_{1:T}'|\overline{x}_{0}')}{p_{\theta}(\overline{x}_{1:T}'|\overline{x}_{0}')}\\
\approx&\arg\max\limits_{\delta}\sum_{t=0}^{T-1}\mathbb{E}_{q(\overline{x}_{t+1}'|\overline{x}_{t}')} -\Vert \overline{x}_{t+1}'-\mu_{p_{\theta'}(\overline{x}_{t+1}'|\overline{x}_{t}')}\Vert ^{2} \\ & + \Vert \overline{x}_{t+1}'-\mu_{p_{\theta}(\overline{x}_{t+1}'|\overline{x}_{t}')}\Vert ^{2}.\\
\end{aligned}
\end{equation}

Proof for this approximation is available in \cref{app: forward-ddim} and \cref{proof:1}. It is notable that the  difference between a given datapoint and its mean value of probability function in forward DDIM is in fact the error for the noise predictor $\epsilon_{\theta}$ at time $t$:

\begin{equation}
\label{eq:approxmiation:result}
    \begin{aligned}
    &\Vert \overline{x}_{t+1}'-\mu_{p_{\theta}(\overline{x}_{t+1}'|\overline{x}_{t}')}\Vert ^{2}\\
    \approx &(\frac{\sqrt{1-\alpha_{t}}\sqrt{\alpha_{t+1}}}{\sqrt{\alpha_{t}}}+\sqrt{1-\alpha_{t+1}} )^2\Vert\varepsilon_{t} - \epsilon_{\theta}(\overline{x}_t', t)\Vert^{2}.\\
\end{aligned}
\end{equation}

The details for this approximation are available in~\cref{approx:proof}. For brevity, we omit the coefficient $\frac{\sqrt{1-\alpha_{t}}\sqrt{\alpha_{t+1}}}{\sqrt{\alpha_{t}}}+\sqrt{1-\alpha_{t+1}} $ of the loss function at different times, in line with the common practice in diffusion models~\cite{ho2020denoising, song2020score}. Consequently, our target can be transformed into:


\begin{equation}
\label{eq:final}
    \begin{aligned}
\delta &:= \arg\max\limits_{\delta}D_{\rm{KL}}(p_{\theta'}(\overline{x}_{1:T}'|\overline{x}_{0}')||{p_{\theta}(\overline{x}_{1:T}'|\overline{x}_{0}')})\\
    &\approx \arg\max\limits_{\delta}\mathbb{E}_{t, \varepsilon_{t} \sim \mathcal{N}(0,1)}\underbrace{\Vert\varepsilon_{t} - \epsilon_{\theta}(\overline{x}_{t}', t)\Vert^{2} - \Vert\varepsilon_{t} - \epsilon_{\theta'}(\overline{x}_{t}', t)\Vert^{2}}_{L_{tar}(\theta, \theta',\overline{x}_0', t, \varepsilon_{t})}.
\end{aligned}
\end{equation}


Intuitively, we establish that the KL divergence between the probability distributions of a given sample, considering two distinct models, can be reformulated as the discrepancy in the MSE loss between the noise prediction error of the two models. 



Importantly, our definition of conceptual differences and the provided proof isn't limited to our problem domain. For example, we demonstrate its effectiveness in membership inference attack (MIA), as detailed in~\cref{app:mia}.

\subsection{Optimizing via Monte Carlo Sampling}
\label{alg}

Direct optimization of the target in \cref{eq:final} is not feasible due to the presence of the gradient of expectation. Drawing inspiration from traditional adversarial attacks like PGD~\cite{pgd} and recent work applying adversarial attacks on diffusion models~\cite{advdm}, we utilize the expectation of the gradient to estimate the gradient of the expectation through Monte Carlo Sampling. For each iteration, we sample a time \(t\) and noise \(\varepsilon_{t} \in \mathcal{N}(0,1)\), and compute the gradient of the loss function in \cref{eq:final}. We then perform one step of optimization using this gradient, which can be summarized as:

\begin{equation}
\label{eq:tar_loss}
\begin{aligned}
    & \nabla_{\overline{x}_{0}} \mathbb{E}_{t, \varepsilon_{t} \sim \mathcal{N}(0,1)}  L_{tar}(\theta, \theta',\overline{x}_0, t, \varepsilon_{t}) \\
     \approx & \mathbb{E}_{t, \varepsilon_{t} \sim \mathcal{N}(0,1)}\nabla_{\overline{x}_0} L_{tar}(\theta, \theta',\overline{x}_0, t, \varepsilon_{t}).
\end{aligned}
\end{equation}
Following existing adversarial attacks~\cite{pgd}, we impose an \(L_{2}\) norm constraint on each step. The sample at step \((t+1)\), denoted as  $\overline{x_0}^{(t+1)}$ is derived from the step-wise length \(\alpha\), the current gradient and the sample from the last step $\overline{x_0}^{(t)}$:

\begin{equation}
    \overline{x}_{0}^{(t+1)} = \overline{x_0}^{(t)} + \alpha \frac{ \nabla_{\overline{x}_{0}^{(t)}}L_{tar}(\theta, \theta',\overline{x}_{0}^{(t)}, t, \varepsilon_{t})}{\Vert\nabla_{\overline{x}_{0}^{(t)}}L_{tar}(\theta, \theta',\overline{x}_{0}^{(t)}, t, \varepsilon_{t})\Vert_{2}}.
\end{equation}

Intuitively, our algorithm iteratively updates the input variable such that for each sampled time variable \(t\) and  noise $\varepsilon_{t}$, the estimated noise is much more accurate for the fine-tuned model $\theta'$ compared to the pretrained model $\theta$. The implementation of the whole framework can be found in~\cref{alg:whitebox}.

Notably, a series of models, referred to as latent diffusion models (LDMs), employs both diffusion and denoising processes in a latent space.  Contrasting gradient inversion on these LDMs follows the same paradigm, and we present the algorithm in~\cref{ldm}.

\begin{algorithm}[htbp]
   \caption{Contrasting Gradient Inversion for Diffusion Model (CGI-DM)}
   \label{alg:whitebox}
\begin{algorithmic}
   \STATE {\bfseries Input:} Partial representation $\overline{x}_0$  of data $x_0$, pretrained model parameter $\theta$, fine-tuned model parameter $\theta'$, number of Monte Carlo sampling steps $N$, step-wise  length $\alpha$. 
   \STATE {\bfseries Output:} Recovered sample $\overline{x_0}^{(N)}$
   \STATE Initialize $\overline{x_0}^{(0)} \leftarrow \overline{x_0}$.
   \FOR{$i=0$ {\bfseries to} $N-1$}
   
   \STATE Sample $t \sim \mathcal{U}(1, 1000)$
   \STATE Sample current noise $\varepsilon_{t}\sim \mathcal{N}(0,1)$
    \STATE $\Delta_{\delta^{(i+1)}} \leftarrow \nabla_{\overline{x}_0^{(i)}} L_{tar}(\theta, \theta',\overline{x}_0^{(i)}, t,\varepsilon_{t})$ in \cref{eq:final}
   \STATE $\delta^{(i+1)}\leftarrow\alpha \frac{\Delta_{\delta^{(i+1)}}}{\Vert \Delta_{\delta^{(i+1)}}\Vert_{2}}$

   \STATE Clip $\delta^{(i+1)}$ s.t. $\Vert \overline{x}_0^{(i)}+ \delta^{(i+1)} - \overline{x}_0^{(0)}\Vert_{2}\leq \epsilon$

      \STATE $\overline{x}_0^{(i+1)}\leftarrow \overline{x}_0^{(i)}+ \delta^{(i+1)} $
   \ENDFOR
\end{algorithmic}
\end{algorithm}
\vspace{-0.05in}

%% file: inputs/experiment.tex
\section{Experiments}

\begin{table*}[h]
\resizebox{\textwidth}{!}{%
\begin{tabular}{ccccccccccccc}
\hline
\multicolumn{13}{c}{\cellcolor[HTML]{C0C0C0}\textbf{Style-Driven: WikiArt Dataset}}                                                                                                                                                                                                                                                         \\ \hline
\multicolumn{1}{l|}{}           & \multicolumn{4}{c|}{Dreambooth (w. prior loss)}                                                             & \multicolumn{4}{c|}{Dreambooth (w/o. prior loss)}                                                           & \multicolumn{4}{c}{Lora}                                                              \\ \hline
\multicolumn{1}{l|}{}           & Acc. (C)$\uparrow$ & Acc. (D)$\uparrow$ & AUC (C)$\uparrow$ & \multicolumn{1}{c|}{AUC (D)$\uparrow$} & Acc. (C)$\uparrow$ & Acc. (D)$\uparrow$ & AUC (C)$\uparrow$ & \multicolumn{1}{c|}{AUC (D)$\uparrow$} & Acc. (C)$\uparrow$ & Acc. (D)$\uparrow$ & AUC (C)$\uparrow$ & AUC (D)$\uparrow$ \\
\multicolumn{1}{c|}{Text2img}   & 0.64                & 0.70                & 0.68                & \multicolumn{1}{c|}{0.74}                & 0.67                & 0.72                & 0.69                & \multicolumn{1}{c|}{0.76}                & 0.60                & 0.66                & 0.61                & 0.68                \\
\multicolumn{1}{c|}{inpainting} & 0.67                & 0.71                & 0.71                & \multicolumn{1}{c|}{0.74}                & 0.71                & 0.75                & 0.77                & \multicolumn{1}{c|}{0.82}                & 0.59                & 0.59                & 0.60                & 0.59                \\
\multicolumn{1}{c|}{Img2img}    & 0.82                & 0.86                & 0.89                & \multicolumn{1}{c|}{0.92}                & 0.83                & 0.90                & 0.90                & \multicolumn{1}{c|}{0.94}                & 0.68                & 0.73                & 0.74                & 0.79                \\
\multicolumn{1}{c|}{CGI-DM}     & \textbf{0.90}       & \textbf{0.95}       & \textbf{0.96}       & \multicolumn{1}{c|}{\textbf{0.98}}       & \textbf{0.86}       & \textbf{0.95}       & \textbf{0.94}       & \multicolumn{1}{c|}{\textbf{0.99}}       & \textbf{0.74}       & \textbf{0.80}       & \textbf{0.81}       & \textbf{0.87}       \\ \hline
\multicolumn{13}{c}{\cellcolor[HTML]{C0C0C0}{\color[HTML]{262626} \textbf{Subject-Driven Generation: Dreambooth Dataset}}}                                                                                                                                                                                                                        \\ \hline
\multicolumn{1}{l|}{}           & \multicolumn{4}{c|}{Dreambooth (w. prior loss)}                                                             & \multicolumn{4}{c|}{Dreambooth (w/o. prior loss)}                                                           & \multicolumn{4}{c}{Lora}                                                              \\ \hline
\multicolumn{1}{l|}{}           & Acc. (C)$\uparrow$ & Acc. (D)$\uparrow$ & AUC (C)$\uparrow$ & \multicolumn{1}{c|}{AUC (D)$\uparrow$} & Acc. (C)$\uparrow$ & Acc. (D)$\uparrow$ & AUC (C)$\uparrow$ & \multicolumn{1}{c|}{AUC (D)$\uparrow$} & Acc. (C)$\uparrow$ & Acc. (D)$\uparrow$ & AUC (C)$\uparrow$ & AUC (D)$\uparrow$ \\
\multicolumn{1}{c|}{Text2img}   & 0.65                & 0.71                & 0.68                & \multicolumn{1}{c|}{0.75}                & 0.69                & 0.75                & 0.70                & \multicolumn{1}{c|}{0.79}                & 0.64                & 0.69                & 0.66                & 0.74                \\
\multicolumn{1}{c|}{inpainting} & 0.63                & 0.71                & 0.67                & \multicolumn{1}{c|}{0.76}                & 0.67                & 0.79                & 0.71                & \multicolumn{1}{c|}{0.82}                & 0.62                & 0.64                & 0.64                & 0.68                \\
\multicolumn{1}{c|}{Img2img}    & 0.67                & 0.77                & 0.73                & \multicolumn{1}{c|}{0.82}                & 0.75                & 0.81                & 0.81                & \multicolumn{1}{c|}{0.87}                & 0.65                & 0.75                & 0.68                & 0.76                \\
\multicolumn{1}{c|}{CGI-DM}     & \textbf{0.84}       & \textbf{0.85}       & \textbf{0.90}       & \multicolumn{1}{c|}{\textbf{0.91}}       & \textbf{0.89}       & \textbf{0.88}       & \textbf{0.94}       & \multicolumn{1}{c|}{\textbf{0.94}}       & \textbf{0.76}       & \textbf{0.79}       & \textbf{0.82}       & \textbf{0.86}       \\ \hline
\end{tabular}%
}
\caption{Comparison of CGI-DM and other existing pipelines in copyright authentication  for style-driven generation using WikiArt dataset~\cite{wikiart} and for object-driven generation under Dreambooth dataset~\cite{dreambooth} under different fine-tuning methods. The Monte Carlo sampling steps ($N$) for CGI-DM is fixed to 1000 and the step-wise length ($\alpha$) is fixed to 2 in $L_{2}$ norm. The overall budget is fixed to 70 in $L_{2}$ norm.  We employ block-wise masking with a block size of 4 for removing partial information. The experimental results demonstrate that CGI-DM exhibits superior performance under all scenarios.}
\label{result}
\vspace{-0.15in}
\end{table*}

\begin{table}[t]
\resizebox{\linewidth}{!}{%
\begin{tabular}{ccccc}
\hline
\multicolumn{1}{l}{Diffusion Model Structures} & Acc. (C)$\uparrow$  & Acc. (D)$\uparrow$  & AUC (C)$\uparrow$  & AUC (D)$\uparrow$ \\ \hline
SD(v1.4)             & 0.93                 & 0.96               & 0.97 & 0.98   \\
SD(v1.5)             & 0.97                 & 0.96       &0.97 &    0.99       \\
SD(v2.0)      &0.88	&0.94	&0.91	&0.98\\
AltDiffusion         & 0.92                 & 0.94           &0.98 &     0.99  \\ \hline
\end{tabular}%
}
\caption{Robustness of CGI-DM to different model structures and parameters. Experiments are conducted under Dreambooth (with prior loss) fine-tuning using 5 classes of images from the WikiArt dataset.  All parameters for CGI-DM are set as the default value in~\cref{result}.}

\label{models}
\vspace{-0.20in}
\end{table}

In this section, we apply our proposed method, CGI-DM, for copyright authentication under various few-shot generation methods across different types of Diffusion Models. For style-driven generation, which focuses on capturing the key style of a set of images, we randomly select 20 artists each with 20 images from the WikiArt dataset~\cite{wikiart}. For subject-driven generation, which emphasizes details of a given object, we randomly choose 30 objects from the Dreambooth dataset~\cite{dreambooth}, each consisting of more than 4 images. Half of these images are utilized for training, while the other half remains untrained. This results in 10 images used for training a style and 2-6 images used for training an object, aligning with the recommended number for training in the mentioned fine-tuning methods~\cite{dreambooth, lora}. We adopt the 
aforementioned image selection process as the default setting.

The default model used for training is Stable-Diffusion-Model V1.4\footnote{https://huggingface.co/CompVis/stable-diffusion}. Additionally, we demonstrate the adaptability of our method to various types and versions of diffusion models, different training steps, and a larger number of training images (refer to~\cref{sec:gen} for more details).

As noted in previous studies~\cite{mem1}, the presence of near-duplicate examples across different datasets complicates the differentiation between seen and unseen datasets. To address this, we define near-duplicate examples as those with Clip-similarity~\cite{clip-score} exceeding 0.90, and we take precautions to prevent their inclusion within the WikiArt datasets (refer to~\cref{datasets_compare} for further details).

Our objective is to apply our method CGI-DM to distinctly differentiate between images used for training (membership) and those left untrained (holdout)~\cite{mia-dm1, mia-dm2}. We conduct tests across various fine-tuning scenarios, including direct Dreambooth (with prior loss), Dreambooth (without prior loss)~\cite{dreambooth}, lora~\cite{lora}. Specifics of the training process are outlined in~\cref{train_details}. It is noteworthy that the Dreambooth (without prior loss) scenario holds particular significance, given that the process closely mirrors direct fine-tuning.

We set Monte Carlo sampling steps ($N$) to 1000 and  the step-wise length ($\alpha$) to 2, with an overall updating of 70  within the $L_{2}$ norm by default for our method.  This value (70) approximates the average distance between the partial representation of the images and the original images.  The 
default block size for deriving partial representation is 4. For all experiments in~\cref{sec:compare}, where our method is compared with other approaches, we utilize all 20 classes from the WikiArt Dataset and all 30 
classes from the Dreambooth Dataset. Additionally, for ablation studies and experiments validating the generalization ability of our method in~\cref{sec:gen}, \cref{sec:abl} and~\cref{sec:defense}, we randomly select five classes of images from the WikiArt Dataset.

\subsection{Evaluation Metrics}

As mentioned in \cref{background:copyright}, it is crucial for the metric to be visualizable and understandable by human beings.  Therefore, upon deriving the extracted images and original input images, we calculate the visual similarity between them.  Specifically, following the approach in a prior study \cite{dreambooth}, we employ Clip similarity and Dino similarity, utilizing the feature space of the Clip~\cite{clip-score}\footnote{We use image embeddings of Clip ViT-B/32.} and Dino models~\cite{dino}. These measures have been demonstrated to align closely with human vision~\cite{clip-score, dreambooth}.


Upon obtaining the similarity scores, we determine an optimal threshold to distinguish between membership and holdout data. Subsequently, we calculate the Accuracy (\textbf{Acc.}) and Area Under the ROC Curve (\textbf{AUC})~\cite{mia-dm1, mia-dm2}  to assess the effectiveness of a method in discriminating between the two, thereby evaluating its performance in authenticating copyright. We abbreviate the Acc. and AUC under similarity measured by Clip as Acc. (C) and AUC (C), and the ones under similarity measured by Dino as Acc. (D) and AUC (D).

Notably, for a given fine-tuning method, we utilize a fixed threshold to distinguish membership and holdout among different objects and styles—an approach that accounts for real-world scenarios. Generally, obtaining membership and holdout datasets beforehand, and determining an optimal threshold for each style or object, is impractical. Therefore, our threshold is set independently of different styles or objects, representing a more practical and meaningful scenario.
Results for a specific threshold for each style or object are also presented in \cref{separate_threshold} to ensure a comprehensive discussion.


\subsection{Comparison with Existing Methods}

\label{sec:compare}
While no methods explicitly claim to be used for  authenticating copyright on few-shot generation, both existing image generation~\cite{legalnews, mem1} and inpainting~\cite{ddnm} pipelines exhibit the potential to do so. Particularly, we compare our method with three types of pipelines that could be employed for identifying infringements:

\noindent\textbf{Text-to-Image Generation.} Using the official Text-to-Image pipeline in diffusers\footnote{https://huggingface.co/docs/diffusers/api/pipelines/stable\_diffusion/text2img}, we generate $100\times\rm{Num}$ images for each fine-tuned model using the prompt employed during training. Here, `$\rm{Num}$' represents the total number of images in the membership and holdout datasets. This process is applied individually to each image in both datasets.
    
\noindent\textbf{Image-to-Image Generation.} Leveraging  the official Image-to-Image pipeline in diffusers\footnote{https://huggingface.co/docs/diffusers/api/pipelines/stable\_diffusion/img2img}, we use the training prompt to generate $100\times\rm{Num}$ images for each fine-tuned model. The inference step is set to 50 and the img2img strength is set to 0.7 for each image in the membership or holdout dataset.
    
\noindent\textbf{Inpainting.} Employing the state-of-the-art inpainting pipeline DDNM~\cite{ddnm}, we generate $100\times\rm{Num}$ images by masking the right half of each image. We use the prompt applied during training for inpainting, and the inference step is fixed at 50.

We set the number of generated images per input image to 100.  This ensures that the time cost for both the baseline method and our method remains similar, facilitating a fair comparison (see \cref{app:time-cost} for more details). We define the similarity between the output images and a given image as the highest similarity achieved among all the generated images corresponding to one target image.

The comparison between CGI-DM and other pipelines is presented in~\cref{result}. It is evident that CGI-DM outperforms others significantly in various few-shot generation scenarios across different datasets.

\begin{table}[t]
\resizebox{\columnwidth}{!}{%
\begin{tabular}{ccccc}
\hline

\multicolumn{1}{l}{\# of Training Images} & Acc. (C)$\uparrow$           & Acc. (D)$\uparrow$     & AUC (C)$\uparrow$           & AUC (D)$\uparrow$      \\
\hline
5                    & 0.90                          & 0.90        & 0.93&     0.97             \\
10                   & 0.93                          & 0.96         &0.97 &      0.98           \\
20                   & 0.86                         & 0.94         & 0.92&           0.97      \\
50                   & 0.86 & 0.96 &0.95 & 0.99\\ \hline
\end{tabular}%
}
\caption{Impact of training image number on the performance of CGI-DM on WikiArt dataset. All other parameters are set the same as those in~\cref{models}.}
\label{images}

\end{table}
\subsection{Generalization}
\label{sec:gen}
In this section, we take a step further to test whether our method can be applied to a broader range of scenarios, including different Diffusion model structures, varying numbers of training images, and different numbers of training steps.


\noindent\textbf{Different DMs.} We select different versions of two distinguishable Diffusion Models: Stable Diffusion Model~\cite{rombach2022high} and AltDiffusion~\cite{ye2023altdiffusion}, which are representative of latent-space DMs and multilingual DMs, respectively. We conduct experiments using the following versions of the two models: SDv1.4\footnote{https://huggingface.co/CompVis/stable-diffusion}, SDv1.5\footnote{https://huggingface.co/runwayml/stable-diffusion-v1-5}, SDv2.0\footnote{https://huggingface.co/stabilityai/stable-diffusion-2}, and AltDiffusion\footnote{https://huggingface.co/docs/diffusers/api/pipelines/alt\_diffusion}.

As shown in~\cref{models}, our method consistently maintains high Acc. (above 88\%) and AUC (above 90\%)  scores across different DMs.

\begin{figure}
  \centering
  \begin{subfigure}{0.48\linewidth}
    \includegraphics[width=\linewidth]{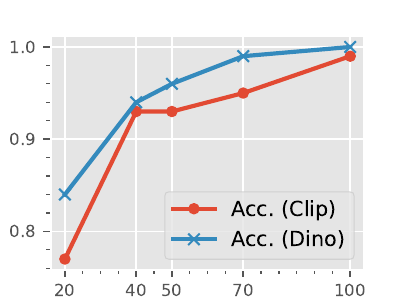}
    \caption{CGI-DM under different training steps}
    \label{training steps}
  \end{subfigure}
  \hfill
  \begin{subfigure}{0.48\linewidth}
    \includegraphics[width=\linewidth]{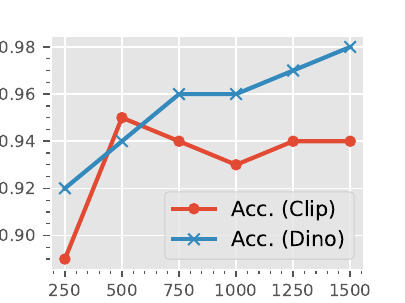}
    \caption{CGI-DM under different extraction steps}
    \label{steps}
  \end{subfigure}
  \vspace{-0.05in}
  \caption{CGI-DM under different training steps and  extraction steps.  All other parameters are set the same as those in~\cref{models}.}
  \label{fig:lines}
\end{figure}
\noindent\textbf{Number of Training Images.} As the number of training images increases, the learned concept during fine-tuning becomes more intricate, thereby enhancing the difficulty of copyright authentication. To thoroughly examine how this influences performance, we conduct experiments with varying numbers of training images, the results of which are presented in \cref{images}. It is evident that our method remains almost stable despite changes in the number of training images. Specifically, CGI-DM consistently achieves 
Acc.~around 90\% in all cases, which demonstrates the robustness of CGI-DM to scenarios with various training data sizes during fine-tuning.



\noindent\textbf{Training Steps.} Increasing training steps is known to lead to stronger 
memorization of the training images, making the extraction process easier~\cite{mem1, mem2}. Therefore, we also analyze the performance of our method under different training steps. As depicted in~\cref{training steps}, our method maintains a high authentication success  rate (above 90\% Acc.) under the Dreambooth scenario (with prior loss), with 40 training steps per image. The 40 training steps per image are notably fewer than the typically 
recommended 100 steps per image for training, as reported in~\cite{dreambooth}. As the number of training steps increases, the model becomes more adept at memorizing the concepts present in the given images, leading to a rise in copyright authentication accuracy for CGI-DM.


\subsection{Ablation Study}
\label{sec:abl}
\subsubsection{Methods for Removing Partial Information}
\label{sec:partial}

~~ As mentioned earlier, our approach commences with a partial representation of the images. In this section, we explore the following techniques to eliminate partial information from the provided images:

\noindent\textbf{Blurring.} We employ Gaussian blurring\footnote{https://pillow.readthedocs.io/en/stable/reference/ImageFilter.html}, with a fixed kernel size of 16 and a blurring rate of 7.

\noindent\textbf{Masking.} We utilize a large mask which covers right half of the images, following previous work~\cite{mem1}.

\noindent\textbf{Block-wise Masking.} We use square with different sizes (2, 4 and 8) to mask half of the region in the images.

\begin{figure}[t]
\vspace{-0.05in}
\begin{center}
\includegraphics[width=\columnwidth]{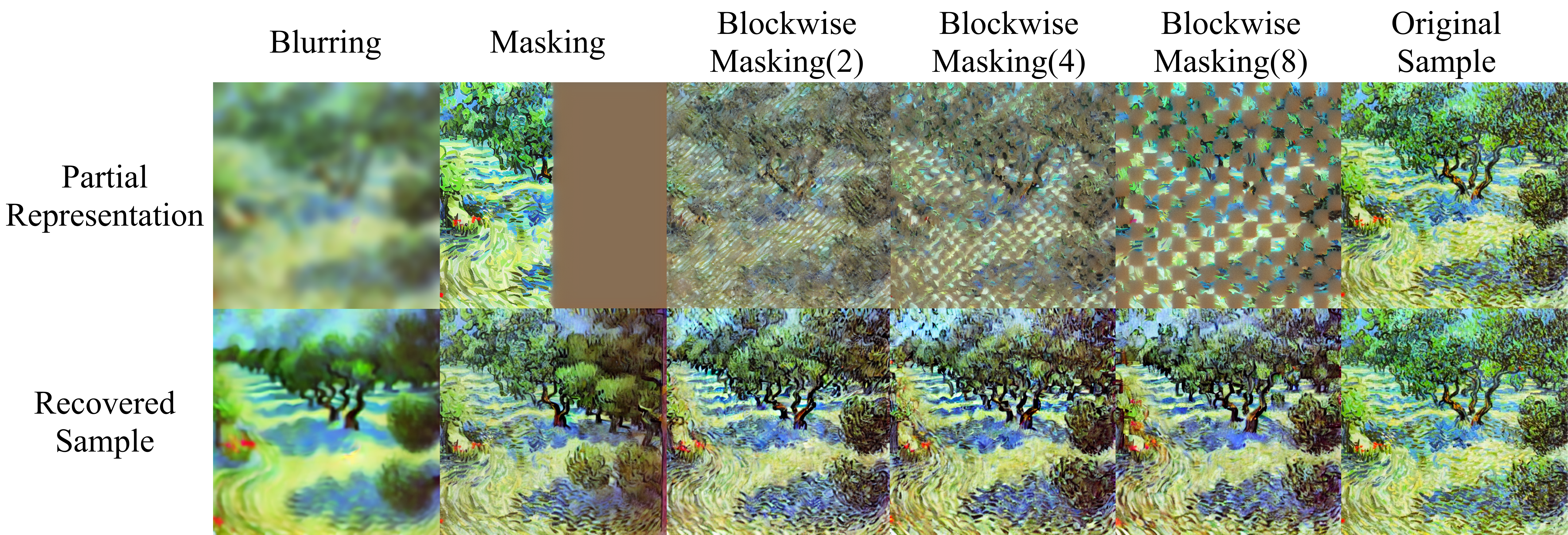}
    \caption{Visualization of different methods for removing partial information and corresponding recovered samples  for CGI-DM on Van Gogh's paintings from the WikiArt dataset.}
    \label{masked}
\end{center}
\vspace{-0.25in}
\end{figure}

The representation is illustrated in \cref{masked}. As indicated in both~\cref{masked} and \cref{masked:tab}, we observe that to achieve a match between the recovered images and the desired training samples, the masked information must be fine-grained, ensuring retention of local information in the masked image. Otherwise, the recovered content may be irrelevant to the given image. Consequently, the block-wise masking emerges as the superior choice. Regarding the block size of the mask, we discover that it does not significantly impact performance.

\begin{table}[t]
\resizebox{\columnwidth}{!}{%
\begin{tabular}{ccccc}
\hline
Removing Methods & Acc. (C)$\uparrow$  & Acc. (D)$\uparrow$ & AUC (C)$\uparrow$  & AUC (D)$\uparrow$  \\ \hline
Blurring                 & 0.65               & 0.80    & 0.69&        0.86     \\
Masking              & 0.72                & 0.73         &0.76 &     0.75   \\
Block-wise Masking(2)    & 0.93 & \textbf{0.97} & \textbf{0.97} & 0.99\\
Block-wise Masking(4)    & 0.93                 & 0.96      &\textbf{0.97} &       0.98     \\
Block-wise Masking(8)    & \textbf{0.94}        & \textbf{0.97}   & \textbf{0.97}&  \textbf{1.00}  \\ \hline
\end{tabular}%
}
\caption{Influence of different approaches for removing partial information on CGI-DM. All other parameters are set as the same~\cref{models}.}
\label{masked:tab}
\end{table}

\subsubsection{Extraction Steps}

~~ One critical hyper-parameter in our algorithm is the number of Monte Carlo Sampling steps, which we denote as ``Extraction steps". Notably, the time required by CGI-DM  increases linearly with the increment in extraction steps. We conduct experiments with different extraction steps and present the results in~\cref{steps}. The findings suggest that excessively small extraction steps fail to yield the optimal solution, leading to inadequate exploitation of the conceptual differences between the fine-tuned models and the pretrained models. Consequently, this inadequacy results in significantly poorer extraction performance and copyright authentication. Notably, performance stabilizes when the number of steps reaches approximately 1000. Further increasing the number of steps substantially escalates the computational cost while yielding only marginal improvements.



\subsection{Performance of CGI-DM under Defenses}
\label{sec:defense}
As highlighted in prior research~\cite{mia-dm1,mia-dm2,mia-whitebox}, it is possible to defend against methods that leverage the loss function of a model. In line with this, we conduct experiments on CGI-DM under various defenses, including RandomHorizontalFlip~\cite{mia-dm1}, Cutout~\cite{mia-defense2}, and RandAugment~\cite{mia-defense3}. The block size for Cutout is consistently fixed at $64 \times 64$. Notably, RandAugment serves as a strong privacy-preserving defense, at the price of a decline in generation quality~\cite{mia-dm1}. The results presented in~\cref{tab:defense} underscore the consistent effectiveness of our method under various defense mechanisms. Even RandAugment, recognized for its strong defense capabilities, demonstrates limited effectiveness in reducing the efficacy of copyright authentication for CGI-DM, with reductions of approximately 10\% in Acc.~and 8\% in AUC.



%% file: inputs/related_work.tex
\begin{table}[t]
\resizebox{\columnwidth}{!}{%
\begin{tabular}{cllll}
\hline
\multicolumn{1}{l}{Defense Methods} & \multicolumn{1}{c}{Acc. (C)$\uparrow$}          & \multicolumn{1}{c}{Acc. (D)$\uparrow$}  & \multicolumn{1}{c}{AUC (D)$\uparrow$}  & \multicolumn{1}{c}{AUC (D)$\uparrow$} \\ \hline
No Defense                          & \multicolumn{1}{c}{0.93} & \multicolumn{1}{c}{0.96}  & 0.97&      0.98 \\
HorizationalFlip~\cite{mia-dm1}                 &   \multicolumn{1}{c}{0.93}                       &            \multicolumn{1}{c}{0.94}     &   0.98 &  0.98     \\
Cutout~\cite{mia-defense2}                       &         \multicolumn{1}{c}{0.87}                 &               \multicolumn{1}{c}{0.93}     & 0.92 &       0.97      \\
RandAugment~\cite{mia-defense3}                         &\multicolumn{1}{c}{0.83}                      &\multicolumn{1}{c}{0.85}    & 0.86 &       0.92             \\ \hline
\end{tabular}%
}
\caption{Performance of CGI-DM under possible defenses. All  parameters are set the same as those in~\cref{models}.}
\label{tab:defense}
\vspace{-0.05in}
\end{table}

\section{Related Work}

\noindent\textbf{Model Inversion (MI).} Model Inversion (MI) Attack was first introduced by Matthew \etal~\cite{gi-source}. The attack focuses on deriving the training dataset of a network based on its parameters, mostly aiming at white-box scenarios~\cite{gi-source, gi1, gi2, gi3, gi4, gi5}. Recent approaches on MI~\cite{gi1, gi2, gi3} highlight the importance of prior information during inversion, emphasizing that the proper prior of a generative model could contribute to the inversion of a classification model. Our inversion from partial representation is also inspired by such findings.

Although model inversion has been extensively researched for classification models over the years, there is no systematic research on model inversion for generative models as far as we know.

\noindent\textbf{Membership Inference Attack (MIA).} Membership Inference Attack~\cite{mia-source, mia-dm1, mia-dm2, mia-whitebox} centers on identifying which  subset of a larger data pool is used for training.  These methods generally leverage the loss function of the model in white-box scenarios where the model parameter is available, or simulate the loss in gray-box or black-box scenarios where the model parameter is not as clear.

Limited research has been conducted on the use of MIA for copyright authentication~\cite{mia-whitebox}, possibly due to its intrinsic limitations. While MIA efficiently assesses the origin of a sample, its reliance on the model's loss function as the metric for classification restricts its visualization capabilities. Consequently, MIA-based techniques typically provide only binary outcomes, hindering their use as strong legal evidence. Further discussion is available in \cref{legal}.

\noindent\textbf{Data Watermark on Diffusion Models.}  Data watermarking~\cite{watermark-dm-1, watermark-dm-2, watermark-dm-3} entails embedding watermarks into training images, enabling post-training extraction from the generated output.  This process holds potential for copyright authentication  by identification of training images via extracting the embedded watermarks in the generated images.  However, prior studies~\cite{watermark-dm-1, watermark-dm-2, watermark-dm-3} show that most methods necessitate a single designated watermark across the entire diffusion model dataset. Efficacy of such an approach in few-shot generation scenarios, where only several input images correspond to a single watermark, remains uncertain.  Moreover, the introduction of watermarks may suffer from image degradation~\cite{watermark-dm-1, watermark-dm-2}, content changes~\cite{watermark-dm-2}, and potential removal~\cite{dm-watermark-removed}.

While recognizing these possible issues, it is crucial to note that watermarks and our proposed method are separate ways to protect copyright, and they do not clash. It is conceivable that in the future, these approaches could converge, combining to create a more robust copyright authentication system.

%% file: inputs/conclusion.tex
\section{Conclusion}

In this paper, we introduce a new method for copyright protection called copyright authentication. Our framework, CGI-DM, validates the use of training samples featuring vivid visual representation, serving as a tool for digital copyright authentication. We start by removing part of the input image. Then, using Monte Carlo sampling and PGD, we exploit the differences between the pretrained and fine-tuned model to recover the removed information. A high similarity between the recovered samples and the original input samples suggests a potential infringement. Through experiments on WikiArt and Dreambooth datasets, we demonstrate CGI-DM's robustness and effectiveness, surpassing alternative approaches. Such experimental results show that CGI-DM is adept at providing legal evidence for art style mimicry and unauthorized image fabrication. In conclusion, CGI-DM not only offers  robust method for infringement validation in the evolving DM landscape but also pioneers the application of gradient inversion in generative models.

%% file: inputs/acknowledgements.tex
\section{Acknowledgements}

We extend our deepest gratitude to Professor Ge Zheng of the Koguan Law School at Shanghai Jiao Tong University for his invaluable insights into legal matters. Our thanks also go to Yiming Xue for offering valuable suggestions on legal issue revisions. Furthermore, we are grateful to Chen Gong and Peishen Yan from Shanghai Jiao Tong University, as well as Qingsi Lai from Peking University, for their substantial contributions and thoughtful feedback.

%% file: inputs/appendix.tex
\clearpage
\setcounter{page}{1}
\maketitlesupplementary
\section{Legal Issues}
\label{legal}
\subsection{Copyright Infringement for Artificial Intelligence-generated Content (AIGC)}

In traditional copyright infringement cases, the initial process of determining whether an image has been ``referenced" in the creation of another image is relatively straightforward. However, the subsequent legal judgment, which involves quantifying the extent of similarity and deciding if it amounts to infringement, presents more significant challenges. This necessitates an examination of substantial similarity aspects, including style and composition, while also considering the originality of the work in question and any authorized use. This stage poses challenges due to the requirement for a nuanced understanding of the intricate aspects of copyright law.

In the traditional context, particularly in the field of painting, the acceptance of visual and comprehensible evidence as a key element for legal decisions has evolved over time. Once established, a high degree of similarity found in such evidence often led to findings of infringement. For instance, an artist was fined a huge amount of money for plagiarizing artworks\footnote{https://www.caixinglobal.com/2023-10-25/chinese-artist-fined-5-million-yuan-in-landmark-case-for-plagiarizing-foreign-work-102120348.html}. In this case, visually similar paintings serve as evidence of infringement, underscoring the significance of visual proof in traditional copyright cases. This sets a precedent, highlighting the need for adapting similar methodologies to address the unique challenges in the AIGC realm.

In the era of AIGC, the stark absence of widely-accepted factual evidence, in contrast to traditional contexts, is evident. The intricate nature of AI algorithms complicates the task of establishing direct references. This has led to a significant shift in legal regulations, moving from focusing primarily on direct infringement judgments to tackling anti-unfair competition concerns. This transition is a response to the lack of clear, visual, and easily interpretable evidence, reflecting the adjustments current regulations are undergoing to accommodate the nuances of AIGC\footnote{https://digichina.stanford.edu/work/how-will-chinas-generative-ai-regulations-shape-the-future-a-digichina-forum/}.

Addressing these complexities in AIGC demands the cultivation of a widely-recognized factual understanding of what constitutes a ``reference." Here, the importance of robust copyright authentication tools becomes apparent. These tools are crucial for not only establishing facts but also for making them visually comprehensible and publicly acceptable. The recent lawsuit involving ``Stable Diffusion"\footnote{https://www.hollywoodreporter.com/business/business-news/artists-copyright-infringement-case-ai-art-generators-1235632929} brings to light the current reliance on visual evidence in copyright lawsuits. It also, however, points to the inadequacy of these methods due to the limited similarity between AI-generated output images and the original training data. Our work with CGI-DM focuses on overcoming this hurdle, offering a method to aid in substantiating infringement claims with higher visual similarity.

In the short term, these copyright authentication tools offer valuable assistance in current lawsuits against AIGC. While they may not be crucial at this juncture, their role in establishing necessary visual and comprehensible evidence is significant as we adapt our legal frameworks to the challenges posed by AIGC. In the long term, their importance grows, as establishing well-accepted facts becomes paramount for the effective adjudication of AIGC-related copyright issues. Tools like the proposed CGI-DM method are increasingly vital in this extended context, which enable a shift in focus from merely contending with anti-unfair competition to directly addressing copyright infringement. This reorientation is vital for ensuring stronger protection of the intellectual property rights of human artists, safeguarding their work against the advancing tide of AIGC.

In conclusion, the development and acceptance of robust validation tools are essential in shaping the future of copyright law enforcement in the context of AIGC. Reflecting on the historical trajectory of how visual evidence gradually became integral to legal judgment in traditional contexts, we see a clear need for a similar evolution in the AIGC landscape. This historical perspective reinforces the importance of societal acceptance in the legal adjudication process and underscores the necessity of adapting these principles to the emerging challenges of AIGC.

\begin{table}[t]
\resizebox{\linewidth}{!}{%
\begin{tabular}{ccccc}
\hline
\multicolumn{1}{l}{}     & Need Preprocessing? & Defense       & Accuracy      & Visualizability    \\ \hline
MIA~\cite{mia-dm1, mia-dm2}                       & \textbf{\XSolidBrush} & \textbf{Hard} & \textbf{High} & Poor              \\
Data Watermark~\cite{watermark-dm-1,watermark-dm-2,watermark-dm-3}           & \Checkmark       & Uncertain     & Uncertain     & Moderate \\
Copyright Authentication & \textbf{\XSolidBrush} & \textbf{Hard} & \textbf{High} & \textbf{Strong}   \\ \hline
\end{tabular}%
}

\caption{Comparative Overview of Post-cautionary Copyright Protection Methods. Bold text highlights advantageous features, while plain text indicates areas where these features are less pronounced.}
\label{comparison-post}
\end{table}

\subsection{More Discussion of Post-Cautionary Methods for Copyright Protection}
This section presents an in-depth analysis of various post-caution methods for copyright protection. The summary table in \cref{comparison-post} illustrates that, in terms of preprocessing, defense, accuracy, and visualizability, copyright authentication emerges as a superior choice.

\noindent\textbf{Membership Inference Attack(MIA).} There are only a limited number of studies that explore the use of MIA for infringement validation~\cite{mia-whitebox}, possibly due to  the inherent limitations of this method. MIA primarily assesses the probability of a sample being part of the training dataset, but its dependency on the model's loss function results in outcomes that are challenging to visualize. This limitation restricts MIA’s use in legal settings, as it provides only binary `yes' or `no' outputs, lacking the detailed nuance needed for legal evidence. Additionally, these binary outcomes, being dependent on the model's loss function, deviate from human visual perception, thus limiting their reliability and applicability in the process of legal judgment.


\noindent\textbf{Data Watermark.} Data watermarking typically requires preprocessing on training images~\cite{watermark-dm-1,watermark-dm-2,watermark-dm-3}. This method's limitations include potential degradation in image quality and the feasibility of watermark removal. Furthermore, its effectiveness is predominantly aligned with current Digital Models (DMs) and may not extend to future DMs or other generative models. The intrinsic design of data watermarking implies permanent security once implemented. Besides, its performance under defense and its accuracy remain uncertain under few-shot generation scenarios. The method offers moderate visualizability since the embedded watermark can be discerned in the output, making it more comprehensible than MIA.

\noindent\textbf{Copyright Authentication.} At present, copyright authentication requires no preprocessing and demonstrates robustness in defense scenarios, achieving high accuracy and strong visualizability. These attributes make it an effective tool for providing legal support in cases of copyright infringement.

It is noteworthy that the methods aforementioned do not mutually exclude each other. In the long term, they could potentially converge to form a comprehensive system for validating copyright infringement.

\section{Proof Details}
\label{proof}
\subsection{Forward DDIM}
 \label{app: forward-ddim}
It has recently been found that the diffusion process from \(x_{T}\) to \(x_{1} \) can be directly obtained by leveraging \(x_{0}\) \cite{DDIM}. The reverse of this formulation can even lead to a predicted forward process using the model weight of the diffusion model \cite{diffusion-clip}:

\begin{equation}
\label{eq:foward-DDIM}
\begin{aligned}
  p_{\theta}(x_{t+1}|x_{t}) = & \mathcal{N}(x_{t+1}; \sqrt{\alpha_{t+1}}f_{\theta}(x_{t}, t) +\\& \sqrt{1-\alpha_{t+1}}\epsilon_{\theta}(x_{t}, t), \sigma_{t}^{2} \textbf{I}),  
\end{aligned}
\end{equation}
where $f_{\theta}(x_{t}, t)$ is the predicted $x_{0}$ defined as:$f_{\theta}(x_{t}, t):=\frac{x_{t}-\sqrt{1-\alpha_{t}}\epsilon_{\theta}(x_{t}, t)}{\sqrt{\alpha_{t}}}$.

\subsection{Markov Chain Expanding and Expectation Term Transformation}
\label{proof:1}

KL divergence $D_{\rm{KL}}(p_{\theta'}(\overline{x}_{1:T}'|\overline{x}_{0}')||{p_{\theta}(\overline{x}_{1:T}'|\overline{x}_{0}')})$in \cref{eq:definition} can be extended to the expectation of the logarithm of the division of two probability functions, given the pretrained model and the fine-tuned model.

\begin{equation}
\label{eq:realtarget}
\begin{aligned}
    \delta &:= \arg\max\limits_{\delta}\mathbb{E}_{p_{\theta'}(x'_{1:T}|x'_{0})}\log \frac{p_{\theta'}(x'_{1:T}|x'_{0})}{p_{\theta}(x'_{1:T}|x'_{0})},\\
    &\mbox{where } x_0\sim q(x_0), x_0'=x_0+\delta.
\end{aligned}
\end{equation}

We isolate a single term within the logarithm to simplify the notation and defer the division of two probability terms to a later step. 
\begin{equation}
\label{eq:transformation}
\begin{aligned}
    & \max\limits_{\delta}\mathbb{E}_{p_{\theta'}(x'_{1:T}|x'_{0})}\log p_{\theta'}(x'_{1:T}|x'_{0}) \\
    =& \max\limits_{\delta}\mathbb{E}_{\prod_{t=0}^{T-1} p_{\theta'}(x'_{t+1}|x'_{t})}\log \prod_{t=0}^{T-1} p_{\theta'}(x'_{t+1}|x'_{t})\\
    =& \max\limits_{\delta}\sum_{t=0}^{T-1}\mathbb{E}_{\prod_{k=0}^{T-1} p_{\theta'}(x'_{k+1}|x'_{k})}\log p_{\theta'}(x'_{t+1}|x'_{t})\\
    =& \max\limits_{\delta}\sum_{t=0}^{T-1}\mathbb{E}_{p_{\theta'}(x'_{t+1}|x'_{t})}\log p_{\theta'}(x'_{t+1}|x'_{t}).
\end{aligned}
\end{equation}

The last equation holds true as the possibility function \(p(x'_{t+1}|x'_{t})\) is a Markov chain and is therefore independent of any possibility that precedes time \(t\). It is important to note that, for a well-trained model on data points \(x_{0}\sim q(x_0)\), it should effectively simulate the real prior distribution. In other words, \(p_{\theta'}(x'_{t+1}|x'_{t})\) can be approximated as \(q(x'_{t+1}|x'_{t})\). Notably, this approximation only holds for the fine-tuned model \(\theta'\) as it is the only model trained on image \(x_0\), and it does not hold true for the pretrained model \(\theta\). Based on this approximation, we can modify our target to:


\begin{equation}
\label{eq:appromixation:q}
\begin{aligned}
&\max\limits_{\delta}\sum_{t=0}^{T-1}\mathbb{E}_{p_{\theta'}(x'_{t+1}|x'_{t})}\log p_{\theta'}(x'_{t+1}|x'_{t})\\
    \approx & \max\limits_{\delta}\sum_{t=0}^{T-1}\mathbb{E}_{q(x'_{t+1}|x'_{t})}\log p_{\theta'}(x'_{t+1}|x'_{t}).
\end{aligned}
\end{equation}

\subsection{Approximation of the loss function}
\label{approx:proof}

Given Equation~(\ref{eq:foward-DDIM}) in Section~\ref{background}, we can demonstrate that the discrepancy between the initial image $x_{0}$ and its predicted counterpart $x_{0}:=f_{\theta}(x_{t}, t)$ is indeed linked to the disparity between the forecasted noise $\epsilon_{\theta}(x_t, t)$ at time $t$ with respect to the true noise $\varepsilon_{t}$:
\begin{equation}
\label{eq:preliminary}
\begin{aligned}
& x_{0} - f_{\theta}(x_{t}, t)\\
=& x_{0} - \frac{x_{t} - \sqrt{1-\alpha_{t}}\epsilon_{\theta}(x_{t}, t)}{\sqrt{\alpha_{t}}}\\
=& x_{0} - \frac{\sqrt{\alpha_{t}}x_{0} + \sqrt{1-\alpha_{t}}\varepsilon_{t} - \sqrt{1-\alpha_{t}}\epsilon_{\theta}(x_{t}, t)}{\sqrt{\alpha_{t}}}\\
=& \frac{\sqrt{1-\alpha_{t}}}{\sqrt{\alpha_{t}}}(\varepsilon_{t} - \epsilon_{\theta}(x_t, t)).
\end{aligned}
\end{equation}

Based on this finding, we are now able to establish the validity of the approximation presented in Equation~(\ref{eq:approxmiation:result}):

\begin{equation}
\label{eq:appromixation:proof}
\begin{aligned}
& \Vert x'_{t+1}-\mu_{p_{\theta}(x'_{t+1}|x'_{t})}\Vert ^{2}\\
=& \Vert \sqrt{\frac{\alpha_{t+1}}{\alpha_{t}}} x'_{t} + \sqrt{1-\frac{\alpha_{t+1}}{\alpha_{t}}}\varepsilon_{t+1} - \sqrt{\alpha_{t+1}}f_{\theta}(x'_{t}, t) - \\ &  \sqrt{1-\alpha_{t+1}}\epsilon_{\theta}(x'_{t}, t)\Vert^{2}\\
=& \Vert \sqrt{\alpha_{t+1}} x'_{0} + \sqrt{(1-\alpha_{t})\frac{\alpha_{t+1}}{\alpha_{t}}}\varepsilon_{t} + \sqrt{1 \frac{\alpha_{t+1}}{\alpha_{t}}}\varepsilon_{t+1} - \\&  \sqrt{\alpha_{t+1}}f_{\theta}(x'_{t}, t) - \sqrt{1-\alpha_{t+1}}\epsilon_{\theta}(x'_{t}, t)\Vert^{2}\\
=&\Vert \sqrt{\alpha_{t+1}} (x'_{0}-f_{\theta}(x'_{t}, t)) - \sqrt{1-\alpha_{t+1}}(\epsilon_{\theta}(x'_{t}, t)-\varepsilon_{t}) + \\&(\sqrt{(1-\alpha_{t})\frac{\alpha_{t+1}}{\alpha_{t}}}-\sqrt{1-\alpha_{t+1}})\varepsilon_{t} + \sqrt{1 - \frac{\alpha_{t+1}}{\alpha_{t}}}\varepsilon_{t+1}
\Vert^{2} \\
\approx& \Vert \sqrt{\alpha_{t+1}} (x'_{0}-f_{\theta}(x'_{t}, t)) - \sqrt{1-\alpha_{t+1}}(\epsilon_{\theta}(x'_{t}, t)-\varepsilon_{t})\Vert^{2}.
\end{aligned}
\end{equation}

The final equation is valid due to the approximations of both $\sqrt{1 - \frac{\alpha_{t+1}}{\alpha_{t}}}$ and $(\sqrt{(1-\alpha_{t})\frac{\alpha_{t+1}}{\alpha_{t}}}-\sqrt{1-\alpha_{t+1}})$ being close to 0. Hence, we obtain:

\begin{equation}
\label{eq:approximation:proof2}
\begin{aligned}
&\Vert x'_{t+1}-\mu_{p_{\theta}(x'_{t+1}|x'_{t})}\Vert ^{2}\\\approx&  \Vert \sqrt{\alpha_{t+1}} (x'_{0}-f_{\theta}(x'_{t}, t)) - \sqrt{1-\alpha_{t+1}}(\epsilon_{\theta}(x'_{t}, t)-\varepsilon_{t})\Vert^{2}\\
=&-(\frac{\sqrt{1-\alpha_{t}}\sqrt{\alpha_{t+1}}}{\sqrt{\alpha_{t}}}+\sqrt{1-\alpha_{t+1}} )^{2}\Vert\varepsilon_{t} - \epsilon_{\theta}(x_t, t)\Vert^{2} 
\end{aligned}
\end{equation}

\section{Direct Gradient Inversion (GI) Leads to Zero-information Extracted Result}
\label{direct-gi}
We observe that inverting the gradient solely based on the fine-tuned model parameterized by $\theta'$ resulted in a zero-information-extracted outcome. Specifically, for the partial representation $\overline{x_{0}}$ of an image $x_{0}$, we optimize it based on the following target:
\begin{equation}
\begin{aligned}
            x'_{0}&=\arg\min\limits_{x'_{0}}p_{\theta'}(x'_{1:T}|x'_{0})\\&\approx \arg\min\limits_{x'_{0}}\mathbb{E}_{t, \epsilon_{t} \sim \mathcal{N}(0,1)}  \Vert\varepsilon_{t} - \epsilon_{\theta'}(x'_{t}, t)\Vert^{2}.
\end{aligned}
\end{equation}

\begin{figure}[t]
\vspace{-0.3in}
\begin{center}
\includegraphics[width=\columnwidth]{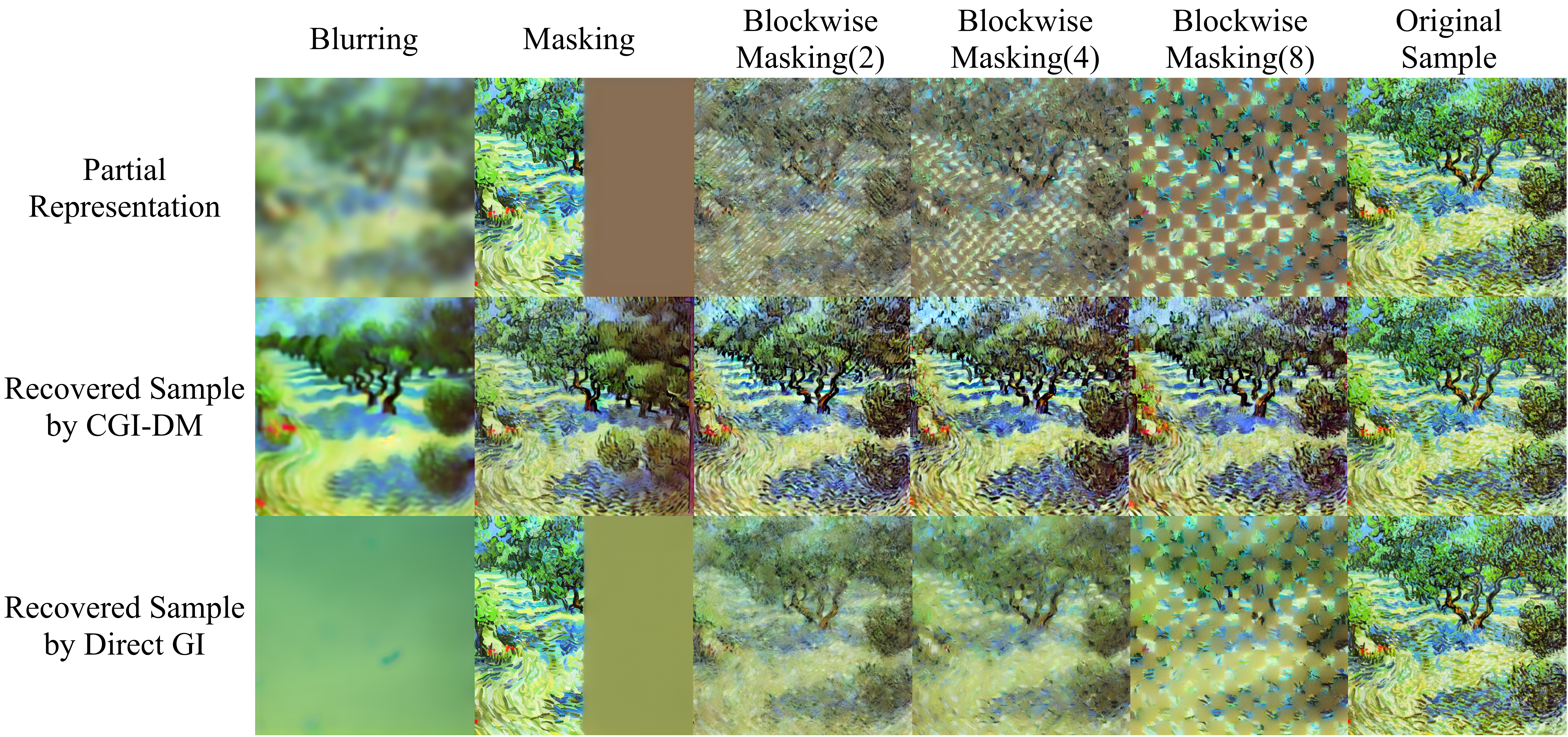}
    \caption{Comparison of CGI-DM and direct GI under different partial representations. We can find that direct GI does not recover any semantic information of the images.}
    \label{nosemantic}
\end{center}
\vspace{-0.4in}
\end{figure}

We omit the proof, as it closely resembles the proof mentioned in Section~\ref{kl}. We employ a similar optimization method utilizing Monte Carlo Sampling and PGD attack, as discussed in Section~\ref{alg}. As illustrated in Figure~\ref{nosemantic}, the extracted images appear blurred, with no useful information discernible. This is primarily attributed to the fact that the diffusion model functions as a robust noise predictor, resulting in the most probable image according to the model being a smooth one with low information content, thus enabling the model to easily predict the introduced noise.

\section{CGI-DM on Latent Diffusion Model (LDM)}
\label{ldm}
The key distinction between the latent diffusion model (LDM) model~\cite{rombach2022high} and others lies in the occurrence of the diffusion process within the latent space. Consequently, our algorithm, CGI-DM, for LDM, closely resembles CGI-DM for DM, with the exception that the partial information removal and the consideration of predicted noise differences are both based on the latent space of a LDM.  Refer to Algorithm~\ref{alg:ldm} for more details.





\begin{algorithm}[htbp]
   \caption{CGI-DM for  LDMs}
   \label{alg:ldm}
\begin{algorithmic}
   \STATE {\bfseries Input:} Partial representation $\overline{x}_0$  of data $x_0$, pretrained model parameter $\theta$, fine-tuned model parameter $\theta'$, number of Monte Carlo sampling steps $N$, step-wise  length $\alpha$, encoder $\mathcal{E}$, decoder $\mathcal{D}$
   \STATE {\bfseries Output:}  Recovered sample $\overline{x}_{0}^{(N)}$
   \STATE Initialize $\overline{z_0}^{(0)} \leftarrow \mathcal{E}(\overline{x_0})$.
   \FOR{$i=0$ {\bfseries to} $N-1$}
   
   \STATE Sample $t \sim \mathcal{U}(1, 1000)$
   \STATE Sample current noise $\varepsilon_{t}\sim \mathcal{N}(0,1)$
    \STATE $\Delta_{\delta^{(i+1)}} \leftarrow \nabla_{\overline{z}_0^{(i)}} L_{tar}(\theta, \theta',\overline{z}_0^{(i)}, t,\varepsilon_{t})$ in \cref{eq:final}
   \STATE $\delta^{(i+1)}\leftarrow\alpha \frac{\Delta_{\delta^{(i+1)}}}{\Vert \Delta_{\delta^{(i+1)}}\Vert_{2}}$

   \STATE Clip $\delta^{(i+1)}$ s.t. $\Vert \overline{z}_0^{(i)}+ \delta^{(i+1)} - \overline{z}_0^{(0)}\Vert_{2}\leq \epsilon$

      \STATE $\overline{z}_0^{(i+1)}\leftarrow \overline{z}_0^{(i)}+ \delta^{(i+1)} $
   \ENDFOR
   
   \STATE $\overline{x}_{0}^{(N)} \leftarrow \mathcal{D}(\overline{z}_{0}^{(N)})$.
\end{algorithmic}
\end{algorithm}
\vspace{-0.05in}

\begin{figure*}[t]
  \centering
  \begin{subfigure}{0.48\linewidth}
    \includegraphics[width=\linewidth]{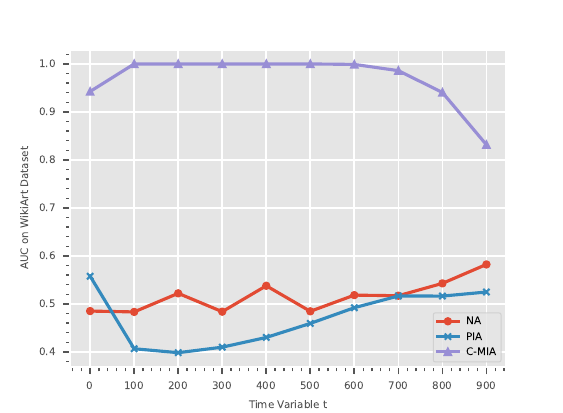}
    \caption{MIA methods under WikiArt Dataset}
    \label{fig:cmia2}
  \end{subfigure}
  \hfill
  \begin{subfigure}{0.48\linewidth}
    \includegraphics[width=\linewidth]{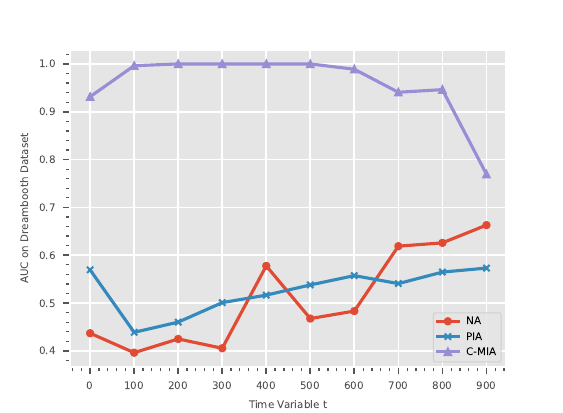}
    \caption{MIA methods under Dreambooth Dataset}
    \label{fig:cmia3}
  \end{subfigure}
  \vspace{-0.05in}
  \caption{Comparison of C-MIA and other MIA methods under two representative datasets under few-shot generation scenarios.}
  \label{fig:cmia1}
\end{figure*}

\section{Removal of near-duplicate samples}
\label{datasets_compare}

Following the precedent set by previous research, we classify samples with a clip similarity exceeding 0.90 as near-duplicates~\cite{mem1}. Within the WikiArt dataset, we eliminate near-duplicate samples until only one copy remains. Given that the utilization of any of these near-duplicate samples already constitutes a violation of the provided data, the presence of such samples in both the training dataset and other datasets would result in an impractically low assessment of the effectiveness of the infringement validation algorithm.

\section{Conceptual Difference Exploiting for MIA (C-MIA)}
\label{app:mia}

The definition of conceptual differences and the provided proof in \cref{kl} isn't limited to copyright authentication, which has strong effects in MIA under few-shot generation scenarios. In detail, for pretrained model $\theta$, fine-tuned mode $\theta'$ and a given sample $x$, we leverage the KL divergence of the probability of $x$ between the two models:

\begin{equation}
\label{eq:c-mia}
    \begin{aligned}
&D_{\rm{KL}}(p_{\theta'}(x_{1:T}|x_{0})||{p_{\theta}(x_{1:T}|x_{0})})\\
    &\approx \mathbb{E}_{t, \varepsilon_{t} \sim \mathcal{N}(0,1)}\underbrace{\Vert\varepsilon_{t} - \epsilon_{\theta}(x_{t}, t)\Vert^{2} - \Vert\varepsilon_{t} - \epsilon_{\theta'}(x_{t}, t)\Vert^{2}}_{L_{tar}(\theta, \theta',x_0, t, \varepsilon_{t})}.
\end{aligned}
\end{equation}

We omit details for this approximation as it mainly follows the provided proof in \cref{proof} and \cref{method-sec}. We then utilize such divergence under one given time $t$ and sampled noise $\varepsilon_{t}$, denoted as $L_{tar}(\theta, \theta',x_0, t, \varepsilon_{t})$, directly for MIA. We name this method as ``C-MIA". We experiment under both WikiArt dataset and Dreambooth dataset, with settings aligned to the one mentioned in \cref{result}.  Our method is compared to two representative MIA methods: Naive Attack (NA)~\cite{mia-dm2} and  Proximal Initialization Attack (PIA)~\cite{mia-dm2}. The performance comparison of these methods is shown in  \cref{fig:cmia1}, where we observe consistent outperformance by C-MIA across various time variable choices and datasets.

\begin{table*}[h]
\resizebox{\textwidth}{!}{%
\begin{tabular}{ccccccccccccc}
\hline
\multicolumn{13}{c}{\cellcolor[HTML]{C0C0C0}\textbf{Style Transferring: Wikiart Dataset}}                                                                                                                                                                                                                                                         \\ \hline
\multicolumn{1}{l|}{}           & \multicolumn{4}{c|}{DreamBooth(w. prior loss)}                                                             & \multicolumn{4}{c|}{DreamBooth(w/o. prior loss)}                                                           & \multicolumn{4}{c}{Lora}                                                              \\ \hline
\multicolumn{1}{l|}{}           & ASR(Clip)$\uparrow$ & ASR(Dino)$\uparrow$ & AUC(Clip)$\uparrow$ & \multicolumn{1}{c|}{AUC(Dino)$\uparrow$} & ASR(Clip)$\uparrow$ & ASR(Dino)$\uparrow$ & AUC(Clip)$\uparrow$ & \multicolumn{1}{c|}{AUC(Dino)$\uparrow$} & ASR(Clip)$\uparrow$ & ASR(Dino)$\uparrow$ & AUC(Clip)$\uparrow$ & AUC(Dino)$\uparrow$ \\
\multicolumn{1}{c|}{Text2img}   & 0.72                & 0.77                & 0.68                & \multicolumn{1}{c|}{0.75}                & 0.74                & 0.80                & 0.71                & \multicolumn{1}{c|}{0.80}                & 0.71                & 0.74                & 0.63                & 0.70                \\
\multicolumn{1}{c|}{inpainting} & 0.73                & 0.78                & 0.72                & \multicolumn{1}{c|}{0.76}                & 0.77                & 0.84                & 0.77                & \multicolumn{1}{c|}{0.85}                & 0.67                & 0.67                & 0.61                & 0.60                \\
\multicolumn{1}{c|}{Img2img}    & 0.87                & 0.94                & 0.88                & \multicolumn{1}{c|}{0.95}                & 0.91                & 0.95                & 0.92                & \multicolumn{1}{c|}{0.97}                & 0.77                & 0.80                & 0.76                & 0.80                \\
\multicolumn{1}{c|}{CGI-DM}     & \textbf{0.96}       & \textbf{0.98}       & \textbf{0.97}       & \multicolumn{1}{c|}{\textbf{0.99}}       & \textbf{0.96}       & \textbf{0.99}       & \textbf{0.97}       & \multicolumn{1}{c|}{\textbf{1.00}}       & \textbf{0.82}       & \textbf{0.90}       & \textbf{0.83}       & \textbf{0.93}       \\ \hline
\multicolumn{13}{c}{\cellcolor[HTML]{C0C0C0}{\color[HTML]{262626} \textbf{Subject-Driven Generation: DreamBooth Dataset}}}                                                                                                                                                                                                                        \\ \hline
\multicolumn{1}{l|}{}           & \multicolumn{4}{c|}{DreamBooth(w. prior loss)}                                                             & \multicolumn{4}{c|}{DreamBooth(w/o. prior loss)}                                                           & \multicolumn{4}{c}{Lora}                                                              \\ \hline
\multicolumn{1}{l|}{}           & ASR(Clip)$\uparrow$ & ASR(Dino)$\uparrow$ & AUC(Clip)$\uparrow$ & \multicolumn{1}{c|}{AUC(Dino)$\uparrow$} & ASR(Clip)$\uparrow$ & ASR(Dino)$\uparrow$ & AUC(Clip)$\uparrow$ & \multicolumn{1}{c|}{AUC(Dino)$\uparrow$} & ASR(Clip)$\uparrow$ & ASR(Dino)$\uparrow$ & AUC(Clip)$\uparrow$ & AUC(Dino)$\uparrow$ \\
\multicolumn{1}{c|}{Text2img}   & 0.81                & 0.89                & 0.72                & \multicolumn{1}{c|}{0.85}                & 0.82                & 0.92                & 0.78                & \multicolumn{1}{c|}{0.90}                & 0.77                & 0.90                & 0.68                & 0.85                \\
\multicolumn{1}{c|}{inpainting} & 0.79                & 0.87                & 0.70                & \multicolumn{1}{c|}{0.83}                & 0.81                & 0.91                & 0.73                & \multicolumn{1}{c|}{0.90}                & 0.76                & 0.84                & 0.65                & 0.77                \\
\multicolumn{1}{c|}{Img2img}    & 0.92                & 0.95                & 0.78                & \multicolumn{1}{c|}{\textbf{0.94}}       & 0.92                & 0.98                & 0.90                & \multicolumn{1}{c|}{0.98}                & 0.81                & \textbf{0.93}       & 0.73                & 0.90                \\
\multicolumn{1}{c|}{CGI-DM}     & \textbf{0.97}       & \textbf{0.96}       & \textbf{0.96}       & \multicolumn{1}{c|}{\textbf{0.94}}       & \textbf{0.96}       & \textbf{0.99}       & \textbf{0.96}       & \multicolumn{1}{c|}{\textbf{1.00}}       & \textbf{0.92}       & \textbf{0.93}       & \textbf{0.90}       & \textbf{0.91}       \\ \hline
\end{tabular}%
}
\caption{Comparison of CGI-DM and other existing pipelines under separate thresholds. All other settings are fixed the same as in  \cref{result}. The experimental results demonstrate that CGI-DM also exhibits superior performance under separate thresholds.}
\label{separate_threshold_tab}

\end{table*}

\begin{table*}[t]
\resizebox{\textwidth}{!}{%
\begin{tabular}{ccccccc}
\hline
           & Inference Steps/output img & Number of output imgs/Input Img & Overall inference steps & Need backward?            & Time costs(min)/Input Img & Overall Time costs(min) \\ \hline
Text2img   & 50                         & 100                             & 5000                    & \XSolidBrush & 10.36                     & 207.30                  \\
Inpainting & 50                         & 100                             & 5000                    & \XSolidBrush & 9.19                      & 183.89                  \\
Img2img    & 35                         & 100                             & 3500                    & \XSolidBrush & 6.45                      & 129.03                  \\
CGI-DM     & 1000                       & 1                               & 1000                    & \Checkmark & 5.68                      & 113.67                  \\ \hline
\end{tabular}%
}

\caption{Time costs comparison of CGI-DM with other methods. We experiment under Dreambooth (with prior loss) for one classes in WikiArt-dataset for three times and report the mean time costs here.}
\label{time-cost}
\end{table*}

\section{Finetuning Details}
\label{train_details}

The details of the parameters in the fine-tuning methods are presented below. We use $\rm{Num}$ to represent the number of images utilized for training

\begin{itemize}
    \item \textbf{Dreambooth (With Prior)}:  We use the training script provided by Diffusers\footnote{https://github.com/huggingface/diffusers/blob/main/examples/dreambooth/\\train\_dreambooth.py\label{dreambooth-file}}. Both the text encoder and the U-Net are fine-tuned during the training process. By default, the number of training steps is set to $50\times\rm{Num}$, with a learning rate of $2 \times 10^{-6}$. The batch size is set to 1, and the number of class images used for computing the prior loss is $50\times\rm{Num}$ by default. The prior loss weight remains fixed at 1.0. For the WikiArt dataset, the instance prompt is ``sks style", and the class prompt is ``art style". For the Dreambooth dataset, the instance prompt is ``sks object", and the class prompt is ``an object".
    
    \item \textbf{Dreambooth (No Prior)}: We use the training script provided by Diffusers\textsuperscript{\ref{dreambooth-file}}. All default parameters remain the same as in the case of Dreambooth (With Prior), except for the exclusion of the prior loss. Notably, this adjustment will lead to training that closely resembles direct fine-tuning.


    \item \textbf{Lora}: We use the training script provided by Diffusers\footnote{https://github.com/huggingface/diffusers/blob/main/examples/dreambooth/\\train\_dreambooth\_lora.py}. All default parameters remain consistent with the case in Dreambooth (No Prior), with the exception of the learning rate and training steps, which are adjusted to $1 \times 10^{-4}$ and to $100\times\rm{Num}$, respectively.
\end{itemize}

\section{Time costs}
\label{app:time-cost}

In this section, we present the time costs for overall validation and per-image validation for both our method and the compared pipelines. As demonstrated in Table \ref{time-cost}, the time costs for the baseline methods are approximately equal to or greater than ours under $100 \times \rm{Num}$ generated images, ensuring a fair comparison where all methods require a similar amount of time.

\begin{figure}[h]

    \includegraphics[width=\linewidth]{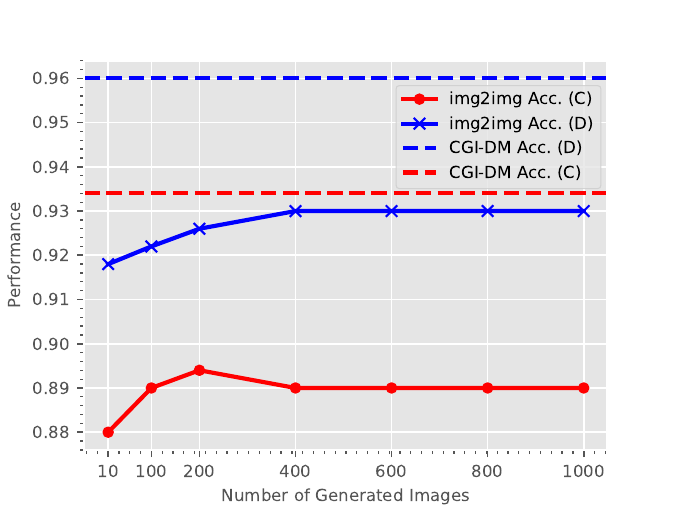}
    \vspace{-0.20in}
    \caption{Trend of img2img baseline. Experimented under 5 subsets of WikiArt dataset.}
    \label{sau}

\end{figure}

We further show the performance of best baseline methods (Img2img) with increasing number of generated images in Table \ref{sau}, where we can observe that $100 \times \rm{Num}$ is near saturated.

\vspace{-0.05in}

\section{Experiment Result under Separate Threshold}
\label{separate_threshold}

The default experiment setting employs a universal threshold across various styles or objects. In this section, we present results obtained with specific thresholds for each style or object. As depicted in \cref{separate_threshold_tab}, CGI-DM also consistently outperforms other methods in this configuration.

\section{Visualization}
\label{sec:visualization}

\begin{figure*}[t]
\vspace{-0.3in}
\begin{center}
\includegraphics[width=\textwidth]{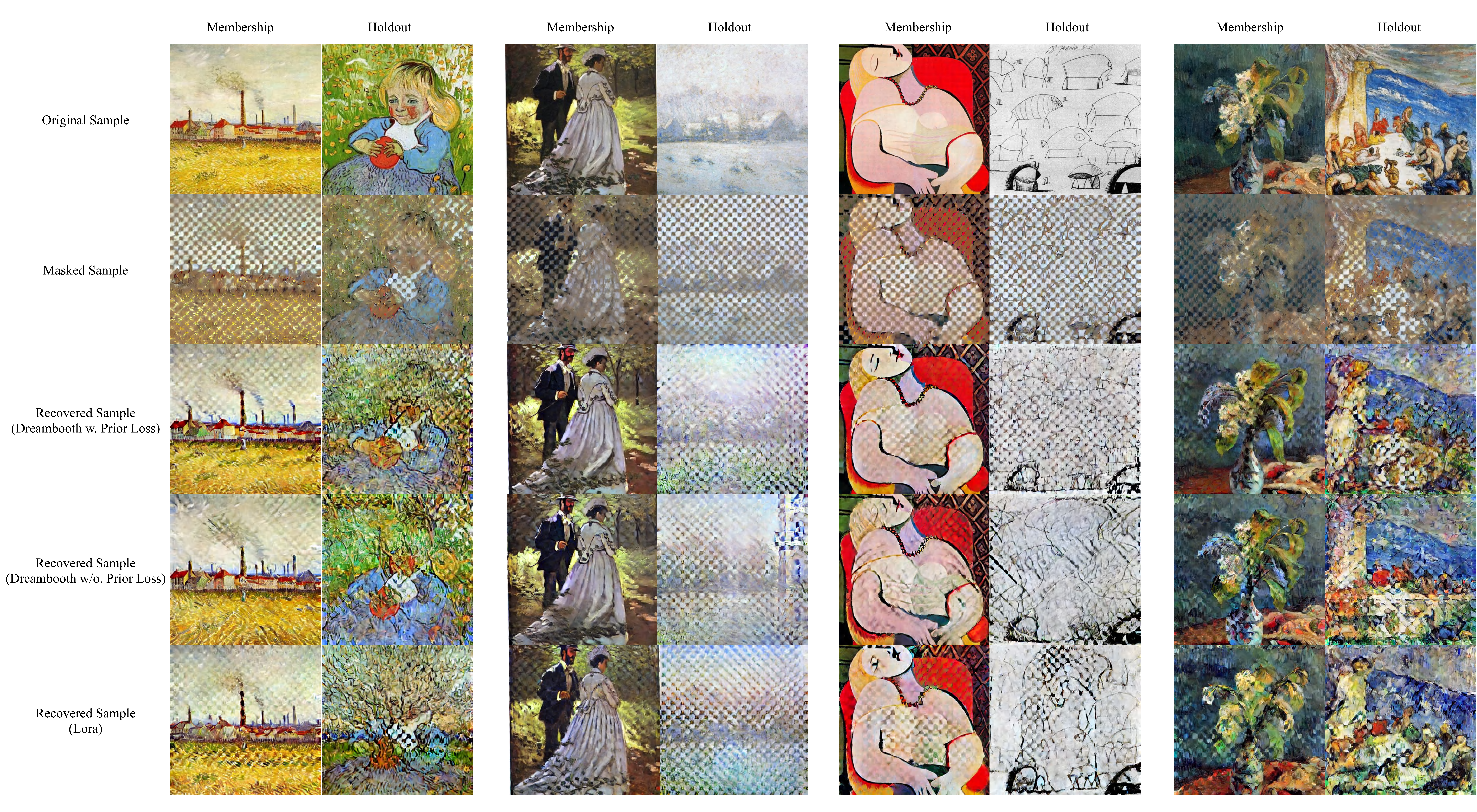}
    \caption{Visualization of CGI-DM for membership and holdout data under 4 classes of WikiArt dataset. From left to right: Vincent Willem van Gogh, Claude Monet, Pablo Ruiz Picasso and Paul Gauguin.}
    \label{vis:WikiArt}
\end{center}

\end{figure*}

\begin{figure*}[t]
\vspace{-0.3in}
\begin{center}
\includegraphics[width=\textwidth]{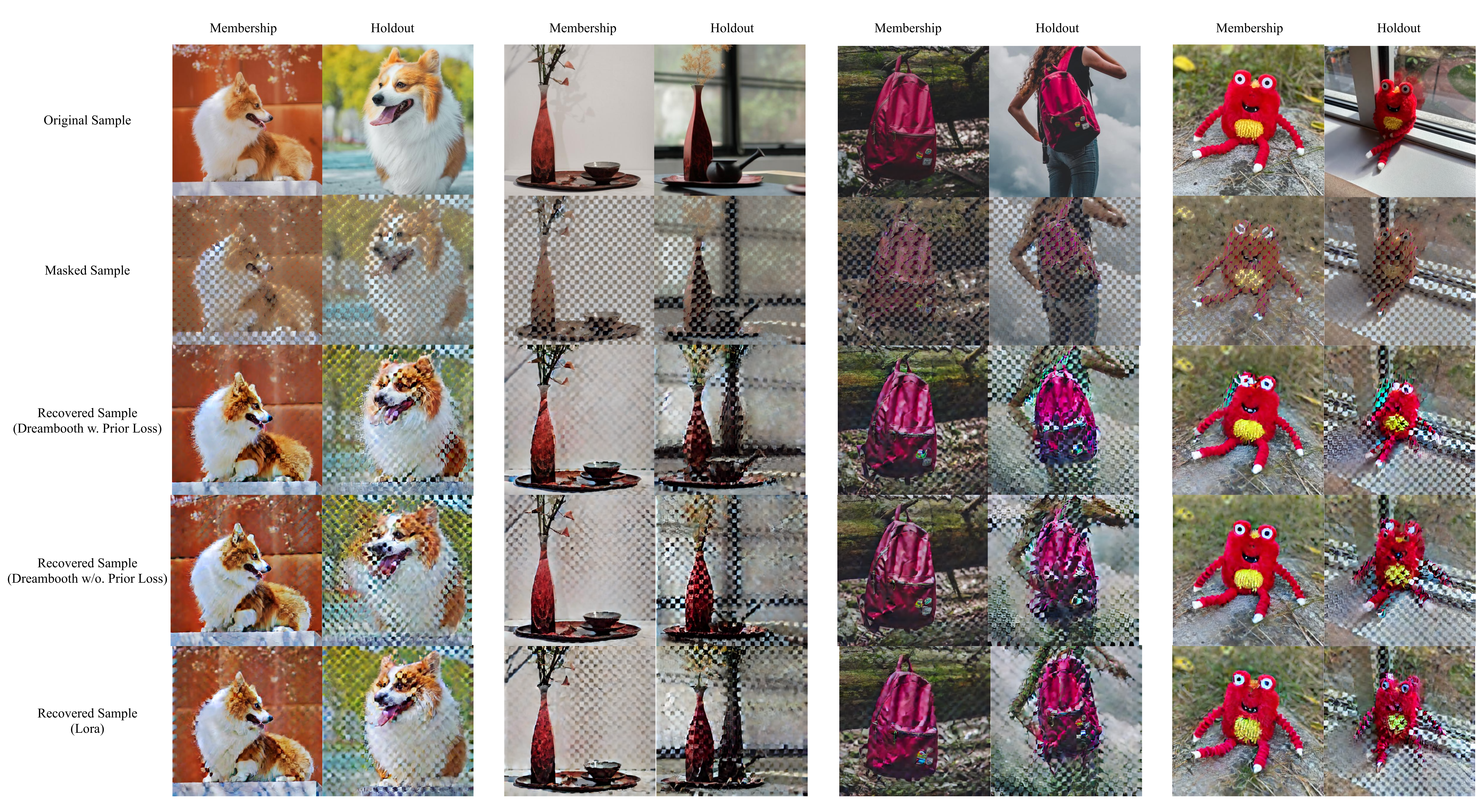}
    \caption{Visualization of CGI-DM for membership and holdout data under 4 classes of Dreambooth dataset. From left to right: dog, vase, backpack and monster\_toy.}
    \label{vis:Dreambooth}
\end{center}

\end{figure*}

\begin{figure*}[t]

\begin{center}
\includegraphics[width=\textwidth]{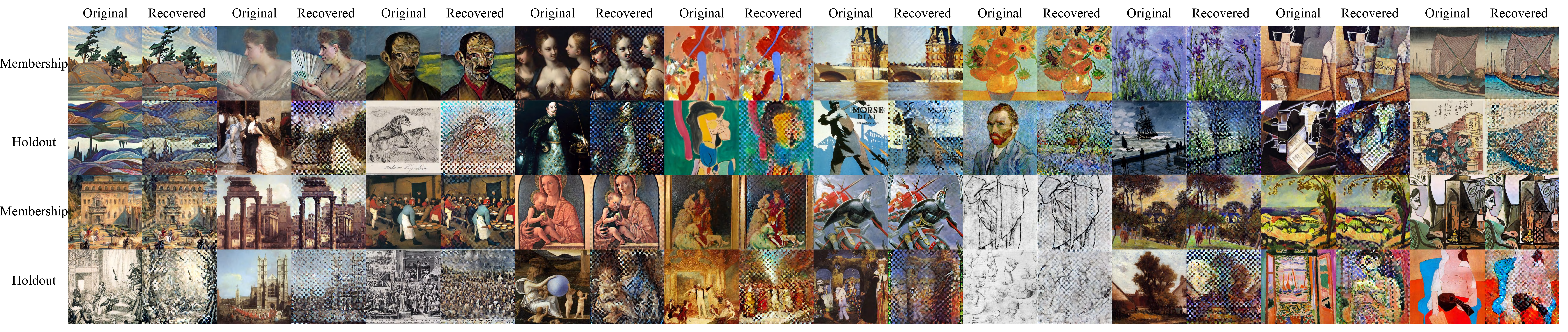}
    \caption{Visualization of CGI-DM for membership and holdout data from 20 different artists in WikiArt dataset under the default setting in  against Dreambooth (with prior loss). The figure is under high resolution and can be zoomed for more details. }
\label{vis:zoom}
\end{center}

\end{figure*}

We present visualization of CGI-DM in \cref{vis:WikiArt}, \cref{vis:Dreambooth} and \cref{vis:zoom}. The experimental setup aligns with the one described in \cref{result}.

\section{No-reference Visual Metrics}
We show results\footnote{\label{myfootnote}Experimental setup matches Tab. 1 in line 402.} in \cref{non-reference} with BRISQUE\footnote{No-Reference Image Quality Assessment in the Spatial Domain (2012).} and CLIP-IQA,\footnote{Exploring CLIP for Assessing the Look and Feel of Images (2022).} where a similar trend can be observed.  It can be seen as a stricter measurement align with human vision and can be helpful for overall evaluations.

\begin{table}[h]
\vspace{-0.10in}
\resizebox{\columnwidth}{!}{%
\begin{tabular}{ccccccc}
\hline
\multicolumn{1}{l}{} & Acc. (C)$\uparrow$ & AUC (C)$\uparrow$ & Acc. (BRISQUE)$\uparrow$ & AUC (BRISQUE)$\uparrow$ & Acc. (CLIP\_IQA)$\uparrow$ & AUC (CLIP\_IQA)$\uparrow$ \\ \hline
CGI-DM               & \textbf{0.90}       & \textbf{0.96}       & \textbf{0.66}          & \textbf{0.69}          & \textbf{0.73}            & \textbf{0.79}            \\
Img2img              & 0.82                & 0.89                & 0.57                   & 0.54                   & 0.52                     & 0.50                     \\
inpainting           & 0.67                & 0.71                & 0.55                   & 0.55                   & 0.52                     & 0.50                   \\
Text2img             & 0.64                & 0.68                & 0.50                   & 0.50                   & 0.50                     & 0.50                     \\ \hline
\end{tabular}%
}
   \vspace{-0.10in}
    \caption{ CGI-DM v.s. baseline under no-reference metrics.}
    \label{non-reference}
    \vspace{-0.15in}
\end{table}

\section{More dicussion with Inpainting Baseline}
The standard approach to inpainting employs half masking techniques. We also delve into inpainting methods that utilize block-wise masking, akin to the CGI-DM process. We include the results in \cref{new}. Experimental setup matches Tab. 1 in line 402.
\begin{table}[h]
\vspace{-0.15in}
\resizebox{\columnwidth}{!}{%
\begin{tabular}{ccccc}
\hline
\multicolumn{1}{l}{}               & Acc. (C)$\uparrow$ & Acc. (D)$\uparrow$ & AUC (C)$\uparrow$ & AUC (D)$\uparrow$ \\ \hline
CGI-DM                             & \textbf{0.90}       & \textbf{0.95}       & \textbf{0.96}       & \textbf{0.98}       \\
inpainting with half masking       & 0.67                & 0.71                & 0.71                & 0.74                \\
inpainting with block-wise masking (4) & 0.71                & 0.80                & 0.77                & 0.87                \\ \hline
\end{tabular}%
}
   \vspace{-0.10in}
    \caption{CGI-DM v.s. inpainting under block-wise masking(4).}
    \label{new}
\vspace{-0.15in}
\end{table}

\newpage

%% file: PaperForReview.bbl
\begin{thebibliography}{10}\itemsep=-1pt

\bibitem{mem1}
Nicolas Carlini, Jamie Hayes, Milad Nasr, Matthew Jagielski, Vikash Sehwag, Florian Tramer, Borja Balle, Daphne Ippolito, and Eric Wallace.
\newblock {Extracting Training Data from Diffusion Models}.
\newblock In {\em USENIX Security}, 2023.

\bibitem{dino}
Mathilde Caron, Hugo Touvron, Ishan Misra, Herv{\'e} J{\'e}gou, Julien Mairal, Piotr Bojanowski, and Armand Joulin.
\newblock {Emerging Properties in Self-supervised Vision Transformers}.
\newblock In {\em ICCV}, 2021.

\bibitem{gi3}
Si Chen, Mostafa Kahla, Ruoxi Jia, and Guo-Jun Qi.
\newblock {Knowledge-enriched Distributional Model Inversion Attacks}.
\newblock In {\em ICCV}, 2021.

\bibitem{mia-defense3}
Ekin~D Cubuk, Barret Zoph, Jonathon Shlens, and Quoc~V Le.
\newblock {Randaugment: Practical Automated Data Augmentation with a Reduced Search Space}.
\newblock In {\em CVPR}, 2020.

\bibitem{watermark-dm-2}
Yingqian Cui, Jie Ren, Han Xu, Pengfei He, Hui Liu, Lichao Sun, and Jiliang Tang.
\newblock {DiffusionShield: A Watermark for Copyright Protection against Generative Diffusion Models}.
\newblock {\em arXiv preprint arXiv:2306.04642}, 2023.

\bibitem{mimicnews}
Andrew Deck.
\newblock {AI-Generated Art Sparks Furious Backlash from Japan’s Anime Community}.
\newblock https://restofworld.org/2022/ai-backlash-anime-artists/, 2022.

\bibitem{mia-defense2}
Terrance DeVries and Graham~W Taylor.
\newblock {Improved Regularization of Convolutional Neural Networks with Cutout}.
\newblock {\em arXiv preprint arXiv:1708.04552}, 2017.

\bibitem{mia-dm1}
Jinhao Duan, Fei Kong, Shiqi Wang, Xiaoshuang Shi, and Kaidi Xu.
\newblock {Are Diffusion Models Vulnerable to Membership Inference Attacks?}
\newblock {\em arXiv preprint arXiv:2302.01316}, 2023.

\bibitem{gi-source}
Matthew Fredrikson, Eric Lantz, Somesh Jha, Simon Lin, David Page, and Thomas Ristenpart.
\newblock {Privacy in Pharmacogenetics: An End-to-End Case Study of Personalized Warfarin dosing}.
\newblock In {\em USENIX Security}, 2014.

\bibitem{gi2}
Ali Hatamizadeh, Hongxu Yin, Pavlo Molchanov, Andriy Myronenko, Wenqi Li, Prerna Dogra, Andrew Feng, Mona~G Flores, Jan Kautz, Daguang Xu, et~al.
\newblock {Do Gradient Inversion Attacks make Federated Learning Unsafe?}
\newblock {\em IEEE Transactions on Medical Imaging}, 42(7):2044--2056, 2023.

\bibitem{clip-score}
Jack Hessel, Ari Holtzman, Maxwell Forbes, Ronan~Le Bras, and Yejin Choi.
\newblock {Clipscore: A Reference-free Evaluation Metric for Image Captioning}.
\newblock {\em arXiv preprint arXiv:2104.08718}, 2021.

\bibitem{ho2020denoising}
Jonathan Ho, Ajay Jain, and Pieter Abbeel.
\newblock {Denoising Diffusion Probabilistic Models}.
\newblock In {\em NeurIPS}, 2020.

\bibitem{lora}
Edward~J Hu, Yelong Shen, Phillip Wallis, Zeyuan Allen-Zhu, Yuanzhi Li, Shean Wang, Lu Wang, and Weizhu Chen.
\newblock {Lora: Low-rank Adaptation of Large Language Models}.
\newblock {\em arXiv preprint arXiv:2106.09685}, 2021.

\bibitem{kawar2022imagic}
Bahjat Kawar, Shiran Zada, Oran Lang, Omer Tov, Huiwen Chang, Tali Dekel, Inbar Mosseri, and Michal Irani.
\newblock {Imagic: Text-Based Real Image Editing With Diffusion Models}.
\newblock {\em arXiv preprint arXiv:2210.09276}, 2022.

\bibitem{diffusion-clip}
Gwanghyun Kim, Taesung Kwon, and Jong~Chul Ye.
\newblock {Diffusionclip: Text-guided Diffusion Models for Robust Image Manipulation}.
\newblock In {\em CVPR}, 2022.

\bibitem{mia-dm2}
Fei Kong, Jinhao Duan, RuiPeng Ma, Hengtao Shen, Xiaofeng Zhu, Xiaoshuang Shi, and Kaidi Xu.
\newblock {An Efficient Membership Inference Attack for the Diffusion Model by Proximal Initialization}.
\newblock {\em arXiv preprint arXiv:2305.18355}, 2023.

\bibitem{advdm}
Chumeng Liang, Xiaoyu Wu, Yang Hua, Jiaru Zhang, Yiming Xue, Tao Song, XUE Zhengui, Ruhui Ma, and Haibing Guan.
\newblock {Adversarial Example Does Good: Preventing Painting Imitation from Diffusion Models via Adversarial Examples}.
\newblock In {\em ICML}, 2023.

\bibitem{pgd}
Aleksander Madry, Aleksandar Makelov, Ludwig Schmidt, Dimitris Tsipras, and Adrian Vladu.
\newblock {Towards Deep Learning Models Resistant to Adversarial Attacks}.
\newblock In {\em ICLR}, 2018.

\bibitem{kimjungginews}
Deborah MT.
\newblock {How AI Art Can Free Artists, Not Replace Them}.
\newblock https://medium.com/thesequence/how-ai-art-can-free-artists-not-replace-them-a23a5cb0461e, 2022.

\bibitem{gi4}
Ngoc-Bao Nguyen, Keshigeyan Chandrasegaran, Milad Abdollahzadeh, and Ngai-Man Cheung.
\newblock {Re-thinking Model Inversion Attacks Against Deep Neural Networks}.
\newblock In {\em CVPR}, 2023.

\bibitem{wikiart}
K. Nichol.
\newblock {Painter by Numbers, WikiArt}.
\newblock https://www.kaggle.com/c/painter-by-numbers, 2016.

\bibitem{mia-whitebox}
Yan Pang, Tianhao Wang, Xuhui Kang, Mengdi Huai, and Yang Zhang.
\newblock {White-box Membership Inference Attacks against Diffusion Models}.
\newblock {\em arXiv preprint arXiv:2308.06405}, 2023.

\bibitem{rombach2022high}
Robin Rombach, Andreas Blattmann, Dominik Lorenz, Patrick Esser, and Bj{\"o}rn Ommer.
\newblock {High-Resolution Image Synthesis with Latent Diffusion Models}.
\newblock In {\em CVPR}, 2022.

\bibitem{dreambooth}
Nataniel Ruiz, Yuanzhen Li, Varun Jampani, Yael Pritch, Michael Rubinstein, and Kfir Aberman.
\newblock {Dreambooth: Fine Tuning Text-to-Image Diffusion Models for Subject-Driven Generation}.
\newblock In {\em CVPR}, 2023.

\bibitem{glaze}
Shawn Shan, Jenna Cryan, Emily Wenger, Haitao Zheng, Rana Hanocka, and Ben~Y Zhao.
\newblock Glaze: Protecting artists from style mimicry by text-to-image models.
\newblock {\em arXiv preprint arXiv:2302.04222}, 2023.

\bibitem{mia-source}
Reza Shokri, Marco Stronati, Congzheng Song, and Vitaly Shmatikov.
\newblock {Membership Inference Attacks against Machine Learning Models}.
\newblock In {\em S\&P}, 2017.

\bibitem{mem2}
Gowthami Somepalli, Vasu Singla, Micah Goldblum, Jonas Geiping, and Tom Goldstein.
\newblock {Understanding and Mitigating Copying in Diffusion Models}.
\newblock {\em arXiv preprint arXiv:2305.20086}, 2023.

\bibitem{DDIM}
Jiaming Song, Chenlin Meng, and Stefano Ermon.
\newblock {Denoising Diffusion Implicit Models}.
\newblock In {\em ICLR}, 2020.

\bibitem{song2020score}
Yang Song, Jascha Sohl-Dickstein, Diederik~P Kingma, Abhishek Kumar, Stefano Ermon, and Ben Poole.
\newblock {Score-Based Generative Modeling Through Stochastic Differential Equations}.
\newblock {\em arXiv preprint arXiv:2011.13456}, 2020.

\bibitem{anti-dreambooth}
Thanh Van~Le, Hao Phung, Thuan~Hoang Nguyen, Quan Dao, Ngoc Tran, and Anh Tran.
\newblock {Anti-DreamBooth: Protecting Users from Personalized Text-to-image Synthesis}.
\newblock In {\em ICCV}, 2023.

\bibitem{legalnews}
James Vincent.
\newblock {The Scary Truth About AI Copyright Is Nobody Knows What Will Happen Next}.
\newblock https://www.theverge.com/23444685/generative-ai-copyright-infringement-legal-fair-use-training-data, 2022.

\bibitem{deepfake:dm}
Tao Wang, Yushu Zhang, Shuren Qi, Ruoyu Zhao, Zhihua Xia, and Jian Weng.
\newblock {Security and Privacy on Generative Data in AIGC: A Survey}.
\newblock {\em arXiv preprint arXiv:2309.09435}, 2023.

\bibitem{ddnm}
Yinhuai Wang, Jiwen Yu, and Jian Zhang.
\newblock {Zero-Shot Image Restoration Using Denoising Diffusion Null-Space Model}.
\newblock In {\em ICLR}, 2022.

\bibitem{watermark-dm-3}
Yuxin Wen, John Kirchenbauer, Jonas Geiping, and Tom Goldstein.
\newblock {Tree-Ring Watermarks: Fingerprints for Diffusion Images that are Invisible and Robust}.
\newblock {\em arXiv preprint arXiv:2305.20030}, 2023.

\bibitem{yang2022diffusion}
Ruihan Yang, Prakhar Srivastava, and Stephan Mandt.
\newblock {Diffusion Probabilistic Modeling for Video Generation}.
\newblock {\em arXiv preprint arXiv:2203.09481}, 2022.

\bibitem{ye2023altdiffusion}
Fulong Ye, Guang Liu, Xinya Wu, and Ledell Wu.
\newblock {AltDiffusion: A Multilingual Text-to-Image Diffusion Model}.
\newblock {\em arXiv preprint arXiv:2308.09991}, 2023.

\bibitem{gi5}
Hongxu Yin, Arun Mallya, Arash Vahdat, Jose~M Alvarez, Jan Kautz, and Pavlo Molchanov.
\newblock {See Through Gradients: Image Batch Recovery via Gradinversion}.
\newblock In {\em CVPR}, 2021.

\bibitem{gi1}
Yuheng Zhang, Ruoxi Jia, Hengzhi Pei, Wenxiao Wang, Bo Li, and Dawn Song.
\newblock {The Secret Revealer: Generative Model-inversion Attacks against Deep Neural Networks}.
\newblock In {\em CVPR}, 2020.

\bibitem{dm-watermark-removed}
Xuandong Zhao, Kexun Zhang, Yu-Xiang Wang, and Lei Li.
\newblock {Generative Autoencoders as Watermark Attackers: Analyses of Vulnerabilities and Threats}.
\newblock {\em arXiv preprint arXiv:2306.01953}, 2023.

\bibitem{watermark-dm-1}
Yunqing Zhao, Tianyu Pang, Chao Du, Xiao Yang, Ngai-Man Cheung, and Min Lin.
\newblock {A Recipe for Watermarking Diffusion Models}.
\newblock {\em arXiv preprint arXiv:2303.10137}, 2023.

\end{thebibliography}
